\documentclass[%
 preprint,
nofootinbib,
 amsmath,amssymb,
 aps,
]{revtex4-1}

\usepackage{mymacros}
\usepackage{graphicx}
\usepackage{dcolumn}
\usepackage{bm}

\begin{document}

\title{Wilsonian Renormalization of Neural Network Gaussian Processes}
\thanks{Alphabetical ordering}

\author{Jessica N. Howard\footnote{Corresponding author}}
\email{jnhoward@kitp.ucsb.edu}
\affiliation{Kavli Institute for Theoretical Physics, Santa Barbara, CA USA}%

\author{Ro Jefferson}%
\email{r.jefferson@uu.nl}
\affiliation{Institute for Theoretical Physics, and Department of Information and Computing Sciences\\Utrecht University, Princetonplein 5, 3584 CC Utrecht, The Netherlands}%

\author{Anindita Maiti}
\email{amaiti@perimeterinstitute.ca}
\affiliation{Perimeter Institute for Theoretical Physics, Waterloo, Ontario, N2L 2Y5, Canada}%

\author{Zohar Ringel}
\email{zohar.ringel@mail.huji.ac.il}
\affiliation{The Racah Institute of Physics, The Hebrew University of Jerusalem, Jerusalem, Israel}%

\date{\today}

\begin{abstract} 
Separating relevant and irrelevant information is key to any modeling process or scientific inquiry. Theoretical physics offers a powerful tool for achieving this in the form of the renormalization group (RG). Here we demonstrate a practical approach to performing Wilsonian RG in the context of Gaussian Process (GP) Regression. We systematically integrate out the unlearnable modes of the GP kernel, thereby obtaining an RG flow of the GP in which the data sets the IR scale. In simple cases, this results in a universal flow of the ridge parameter, which becomes input-dependent in the richer scenario in which non-Gaussianities are included. In addition to being analytically tractable, this approach goes beyond structural analogies between RG and neural networks by providing a natural connection between RG flow and learnable vs. unlearnable modes. Studying such flows may improve our understanding of feature learning in deep neural networks, and enable us to identify potential universality classes in these models.
\end{abstract}

\maketitle


\section{\label{sec:intro} Introduction}

Deep neural networks (DNNs) are showing an increasing amount of universality across tasks \cite{bommasani2022opportunitiesrisksfoundationmodels}, dataset sizes, and computational power \cite{kaplan2020scaling}. In simple asymptotic limits, power-law scaling of DNN performance \cite{kaplan2020scaling} can be traced to Zipf's law behaviour on the underlying datasets  \cite{bahri2021explaining,bordelon2021spectrum}. In other, more non-linear limits, the effects of non-linearities can often be absorbed as a change of a single scale parameter \cite{canatar,LiSompolinsky2021,atanasov2024scaling}. As DNNs are large-scale systems in which the importance of each specific parameter is minute \cite{molchanov2017pruning}, some self-averaging effects leading to universality are perhaps expected, yet their extent requires both theoretical and empirical exploration.  

In physics, scale-free behavior, renormalization of parameters, and universality, are key aspects of the Renormalization Group (RG), one of the pillars of theoretical physics \cite{peskin1995,cardy1996scaling}. At its core, RG is an analytical technique that generates a continuous transformation (flow) between models/physical systems. This transformation is defined such that all models along a given flow agree on low-resolution/coarse-grained observables. Critical theories, showing many of the aforementioned scale-free phenomena, are described by points on this flow at which the correlation length diverges. Besides providing an analytical toolkit for tackling large-scale interacting systems (e.g., \cite{WILSON1974, Gross1973,KT1973}), RG is an inherent part of modelling systems (via notions of relevancy), and offers qualitative insights into the role of non-linearities via the regularities of low-dimensional flows.   

In the context of increasing efforts to make AI safer and more predictable, the aforementioned scale-free universal behaviors of DNNs are encouraging and invite interpretations in terms of some renormalization group flow. At a practical level, there are various  procedures for implementing RG, the choice of which is best determined by physical criteria, chiefly: (i) What is the scale? This determines which observables are being integrated or coarse-grained out, and which remain. (ii) What is the UV theory on which one is applying RG? (iii) Is the induced RG flow tractable, or does it make some quantities more tractable?

An early line of work sought to augment (or interpret) the action of a layer in deep neural networks so that it coarse-grains the data distribution, thus drawing an analogy with RG \cite{mehta2014exact,Koch_Janusz_2018,Beny:2013pmv}. Similarly, the action of a diffusion model can be interpreted as inverting an RG flow on the data distribution \cite{Cotler_2023}. The UV theories here are thus the data distribution and the IR theory consists of the relevant features in the data for the layer action and white noise data for the diffusion action. Another line of work \cite{Erbin_2022,erbin2023functional} 
views the distribution on functions induced by a random DNN as the UV theory, and uses RG formulations to treat non-Gaussian (finite-width) effects \cite{Halverson:2020trp, Grosvenor:2021eol,Roberts_2022} 
in this distribution. However, the notion of scale and the relation to actual DNN priors remains quite varied across these works.  
Another set of works relating RG flows and Bayesian Inference view both as forms of augmenting a probability distribution  \cite{berman2022dynamics,berman2022inverse,Berman_2023,berman2024ncoder,howard2024bayesianrgflowneural}. In this approach however, reducing data generates the RG flow, so the true data-generated model (or target function) is the UV theory, while the  IR theory is a random DNN, so that the observables that are consistently reproduced along the flow are the most unlearnable modes.
To the best of our understanding, these works have yet to analytically predict the generalization performance of neural networks. Finally, an earlier work \cite{Bradde_2017} models the data-distributions, irrespective of any neural network, as a field theory performs RG by removing large PCA components. Trading PCA with kernel-PCA, both that work and the current work share the same notion of UV and IR modes, however focus on very different ensembles/probability distributions.   

Here we formulate a distinct route where RG may come into quantitative use in predicting DNN performance. Our UV theory is the distribution over outputs generated by an ensemble of trained neural networks at finite amount of data, while the IR is the theory for the learnable modes. To generate an RG flow, we integrate out unlearnable output modes and track their renormalizing effect on the learnable modes. To illustrate our approach, we focus on a suitable infinite-overparameterization limit (the Gaussian Process limit) which applies to a wide range of DNNs, including state-of-the-art models such as transformers and Convolutional Neural Networks (CNNs)~\cite{Neal,jacot2018neural,lee2018deep,matthews2018gaussian,Cohen_2021}. We focus on using RG to treat the non-linearities induced by data-averaging/finite-sample-size effects rather than those induced by finite-width, though our formalism naturally extends to that case as well.

We show that in various relevant scenarios, our RG procedure is tractable and results in a renormalized Gaussian Process (GP) theory. When the unlearnable modes are uncorrelated, in a manner to be made precise below, this amounts to a scalar renormalization of the loss or equivalently the ridge parameter, thus providing an RG interpretation for the results of \cite{canatar}; this is developed in sec. \ref{sec:Gaussian}. When correlations between unlearnable modes are present, a further input-dependent/spatial weighting of the loss appears as well as potential shifts to the target; this is demonstrated in sec. \ref{sec:beyond}. The resulting RG flow proceeds until it reaches the first learnable output modes. Thus, our lower RG scale (i.e., the infrared) naturally depends on the type and amount of data. In cases where unlearnable modes have sizable collective fluctuations, the RG flow ends in a weakly interacting regime amenable to perturbative methods \cite{Cohen_2021}. Alternatively, it may end in a strongly interacting theory which, in generic high dimensional scenarios, is amenable to saddle point techniques \cite{canatar}. Lastly, we point out that for power-law decaying kernel spectra, often found in real-world data, we recover scaling-law results \cite{bahri2021explaining}, thereby establishing a concrete link between RG and scaling-laws. 

By relying on a rather general formalism and introducing a down-to-earth practically motivated notion of RG, we hope to provide a stepping stone towards establishing notions of universality in deep learning.

\section{\label{sec:background} Equivalent Kernel Limit of the Replica Action}

To provide a relatively self-contained derivation, we first define Gaussian Process Regression, its relations to neural networks, and replica techniques used to obtain the average predictions over ensembles of datasets. In addition to being a well-established tool in machine learning, Gaussian Process Regression describes both the lazy learning or kernel regime of an infinite-width neural network \cite{lee2018deep,jacot2018neural,Cohen_2021}, alongside extensions to finite networks via approaches detailed in \cite{LiSompolinsky2021,Seroussi2023,ariosto2022statistical}.

The output of such random networks behaves as a Gaussian Process with zero mean and a covariance function $K(x,x')$, called the \emph{kernel}. Here, the training data, $x,x' \in \mathbb{R}^d$, are drawn from a data measure\footnote{By this we simply mean a measure over the data that admits an interpretation as a probability distribution. That is, let $\mathrm{d}\mu_x$ be a measure on $x\in\mathbb{R}^d$ such that for some probability distribution $p_\mathrm{data}(x)$, we have $\int\!f(x)\mathrm{d}\mu_x=\int\! f(x)p_\mathrm{data}(x)\mathrm{d}x$ for $f:\mathbb{R}^d\to\mathbb{R}$. In a slight abuse of notations, we will refer to both the measure $\mathrm{d}\mu_x$ and the distribution $p_\mathrm{data}(x)$ as the data measure.\label{ft:datameasure}} $p_\mathrm{data}(x)$; the total training set consists of $n$ such draws, and is denoted by the $n\times d$ matrix $X_n$. We further consider a regression problem with a target, $y(x)$. A standard result in GP regression states that \cite{Rasmussen} the average predictor on a test point, $\tilde{f}(x_*)$, given $X_n$ is 
\begin{equation}
	\tilde{f}(x_*|X_n) = K(x_*,X_n)[K(X_n,X_n)+\sigma^2 \mathbb{1}_{n\times n}]^{-1} y(X_n)~,
	\label{eq:avgpred}
\end{equation}
where $K(X_n,X_n)$ is understood as an $n \times n$ matrix, and $\sigma^2$ is the variance of the observation noise, called the ridge parameter. (That is, we wish to model more realistic situations in which we do not have access to the function values $y(x)$ themselves, but only to noisy versions thereof, $y(x)+\epsilon$, where $\epsilon$ is additive Gaussian noise with variance $\sigma^2$.) The latter sets an overall tolerance to discrepancies between the target and the prediction on the training set. As we shall see, one main effect of RG is to renormalize this tolerance and also, in more advanced settings (see below), make the ridge parameter input-dependent. 

Since the average predictor \eqref{eq:avgpred} is conditional on $X_n\!\sim\!p_\mathrm{data}(x)$, it is sensitive to the details of the specific dataset we drew. To remove this dependence, we consider the dataset-averaged predictor by computing the expectation value with respect to $p_\mathrm{data}(x)$. The most general way of doing so in a tractable manner uses a ``grand-canonical'' trick \cite{Malzahn, Cohen_2021}, which assumes that the number of training points is drawn from a Poisson distribution with expectation value $\eta$,
\begin{equation}
	\bar{f}(x_*) = \sum_{n=1}^{\infty} \frac{\eta^n e^{-\eta}}{n!}\langle \tilde{f}(x_*|X_n) \rangle_{X_n \sim p_\mathrm{data}(x)} 
	\eqqcolon
	\langle \tilde{f}(x_*|X_n) \rangle_{\eta}~,
	\label{eq:qavg}
\end{equation}
such that $\eta$ is now the average number of data points. Notably, due to the nature of the  Poisson distribution, in the relevant limit of large $\eta$, the fluctuations become increasingly negligible compared to the average.

In statistical physics, \eqref{eq:qavg} is known as a quenched average. Such averages are generally written in terms of the free energy as $\partial_{\alpha} \langle \ln(Z_{\alpha})\rangle_{X \sim p_\mathrm{data}(x)}\big|_{\alpha=0}$, where $Z$ is the partition function and $\alpha$ is some source field that couples to $f(x)$.\footnote{In field-theoretic language, this is just the connected correlation function generated by $\ln Z$. We will suppress the source fields $\alpha$, as they play no role in the rest of the setup.} However, the presence of the logarithm makes explicitly computing this challenging in all but the simplest cases. To this end, a common solution is to use the replica trick, in which we take $M$ copies of the system, and recover $\ln Z$ in the limit $M\to0$ via
\begin{equation}
	\ln Z=\lim_{M\to 0}\frac{1}{M}\!\left(Z^M-1\right)~.
	\label{eq:replica}
\end{equation}
We can now perform the quenched average on the $M$-copy system $Z^M$ instead, which is often more tractable.

This replica approach was first applied to GP regression in \cite{Malzahn}, and was used in \cite{Cohen_2021} to obtain
\begin{equation}
	\langle Z^M \rangle_{\eta} = e^{-\eta}\int\!\prod_{m=1}^{M} \mathcal{D}f_m e^{-S}~,
	\label{eq:repZ}
\end{equation}
where the action is
\begin{equation}
\begin{aligned}
	S=&\sum_{m=1}^M \frac{1}{2}\int\!\mathrm{d}\mu_x\mathrm{d}\mu_{x'}\,f_m(x) K^{-1}(x,x') f_m(x')  - \eta \int\!\mathrm{d}\mu_x\,e^{-\sum_{m=1}^M \frac{\left(f_m(x)-y(x)\right)^2}{2\sigma^2}}~,
	\label{eq:repS}
\end{aligned}
\end{equation}
where $K^{-1}(x,x')$ is the inverse operator (not an element-wise inverse) of $K(x,x')$ having the same eigenfunctions but inverse eigenvalues, and $\mathrm{d}\mu_x$ is the data measure, cf. footnote \ref{ft:datameasure}. As shown in \cite{Cohen_2021} the average predictor $\bar{f}(x_*)$ is given by the average of $M^{-1} \sum\nolimits_{m=1}^M f_{m}(x_*)$ under that action in the limit $M \rightarrow 0$\footnote{To see this, take a derivative of $\left<\left(Z^M-1\right)/M\right>$ with respect to the suppressed source field to obtain $M^{-1}\left<\sum_{m=1}^M f_{m}(x_*)\right>_{Z^M} Z^M$. By replica symmetry (i.e., the fact that the copies are identical), this is equivalent to $\left<f_1(x_*)\right>_{Z^M} Z^M$, which has the same $M\to0$ limit as $\langle f_1(x_*)\rangle_{Z^M}$.}.

Although our initial field theory was Gaussian, averaging over the datasets resulted in a non-quadratic action, due to ``quenched-disorder" induced by the training set. 
A simple yet useful limit \cite{Sollich2004, Silverman1984} of the above theory, known as the Equivalent Kernel (EK) limit, is obtained when $\eta,\sigma^2 \rightarrow \infty$ with $\eta/\sigma^2$ fixed. 
In this case the exponential on the right-hand side of \eqref{eq:repS} can be Taylor expanded\footnote{Note that while the Taylor expansion is exact in this limit, it is still quite a good approximation even at finite $\eta$ and $\sigma^2 = O(0.01)$~\cite{Sollich2004}. It is also a well-known approximation~\cite{Silverman1984} for the finite ridge case.}, so that, at leading order, we obtain a non-interacting theory,
\begin{equation}
\begin{aligned}
	\eta +S =&\sum_{m=1}^M \frac{1}{2}\int\!\mathrm{d}\mu_x \mathrm{d}\mu_{x'}\,f_m(x) K^{-1}(x,x') f_m(x') + \frac{\eta}{\sigma^2} \int\!\mathrm{d}\mu_x \sum_m \frac{\left(f_m(x)-y(x)\right)^2}{2}~.
	\label{eq:quadS}
\end{aligned} 
\end{equation}
It is then convenient to write this in terms of the eigenfunctions, $\phi_k(x)$, of the kernel $K$ with respect to the data measure $\mathrm{d}\mu_x=p_\mathrm{data}(x) dx$. That is, the aforementioned covariance function admits an eigensystem decomposition 
\begin{equation}
	\int\!\mathrm{d}\mu_x\,K(x,x')\phi_k(x)=\lambda_k \phi_k(x')~,
	\label{eq:eigensys}
\end{equation}
where $\phi_k$ is the kernel eigenfunction with eigenvalue $\lambda_k$ which form a complete orthonormal basis, i.e.,
\begin{equation}
	\int\!\mathrm{d}\mu_x\,\phi_k(x)\phi_q(x)=\delta_{kq}~.
	\label{eq:orthophi}
\end{equation}
This allows us to make the following spectral decomposition (analogous to Fourier transforming to momentum space),
\begin{equation}
	f_m(x) = \sum^{\infty}_{k=1} f_{mk} \phi_k(x)~,
	\qquad\quad
	y(x) = \sum^{\infty}_{k=1} y_{k} \phi_k(x)~.
	\label{eq:specdecomp}
\end{equation}
We refer to the coefficients $f_{mk}$ as the \emph{GP modes}, and the functions $\phi_k$ as the \emph{feature modes}. Note that the former carry replica indices and may be highly correlated, while the feature modes are an inert property of the kernel/lazy-network. Inserting these decompositions into the quadratic action \eqref{eq:quadS} yields the decoupled ``feature-space'' action
\begin{equation}
	S = \frac{1}{2}\sum_{m=1}^M \sum_{k=1}^{\infty} \left( \frac{1}{\lambda_k}|f_{mk}|^2 + \frac{\eta}{\sigma^2}|f_{mk}-y_k|^2 \right)~,
	\label{eq:featureS}
\end{equation}
see Appendix \ref{sec:spectral}. Note that the path integral measure also decouples into a measure over the feature modes, $\mathcal{D}f_m\sim \prod_k\mathcal{D}f_{mk}$, up to a constant Jacobian factor.
Given this action, one can then show that the average prediction is \cite{Cohen_2021,Rasmussen}
\begin{equation}\label{eq:naive_EK_avpred}
	\bar{f}_k = \frac{\lambda_k}{\lambda_k + \sigma^2/\eta}\,y_k~,
\end{equation}
with dataset-averaged GP fluctuations given by
\begin{equation}
	\mathrm{Var}[f_k] = \frac{1}{\lambda_k^{-1}+\eta/\sigma^{2}}=\frac{\sigma^2}{\eta} \frac{\lambda_k}{\lambda_k + \sigma^2/\eta}~.
		\label{eq:varfk}
\end{equation}
In the context of our RG prescription introduced below, these expressions have the following important interpretation: modes with $\lambda_k \ll \sigma^2/\eta$ essentially behave as if $\sigma^2 = \infty$ or $\eta=0$, i.e., they are decoupled from the inference problem and hence could just as well have been integrated out. This fact is already implicit in the action \eqref{eq:featureS}, where the term proportional to $\eta/\sigma^2$ becomes negligible compared to $1/\lambda_k$ for sufficiently small $\lambda_k$.

\section{\label{sec:Gaussian} Renormalizing Gaussian features}
In this section, we consider the case in which the distribution of feature modes, $P[\varphi]$ (to be made precise below), is Gaussian. 
This will help establish the notation which we will later use to generalize to the non-Gaussian case in section~\ref{sec:beyond}. 
As we will later see, this Gaussian assumption is commonly made due in part to the fact that the feature distributions of many real-world datasets are (approximately) Gaussian.  
In section~\ref{sec:scalingLaws}, we give a practical demonstration of this effect; namely, we show that our RG approach can predict the scaling of neural network performance with dataset size, matching both empirical observations and previous results obtained with alternative methods~\cite{canatar, bahri2021explaining}. 

Formally, our Wilsonian renormalization procedure can be explained very simply: we wish to integrate out all modes above some cutoff $\kappa$, in order to obtain an effective action in terms of only the modes with $k\leq\kappa$.
Accordingly, we split the spectral decomposition \eqref{eq:specdecomp} into
\begin{equation}
	\GPlow(x)\coloneqq \sum_{k\leq\kappa}f_{mk} \phi_k(x)~,
	\quad
	\GPhigh(x)\coloneqq \sum_{k>\kappa} f_{mk} \phi_k(x)~,
	\label{eq:split1}
\end{equation}
and similarly
\begin{equation}
	y_<(x) \coloneqq \sum_{k\leq\kappa} y_{k} \phi_k(x)~,
	\quad
	y_>(x) \coloneqq \sum_{k>\kappa} y_{k} \phi_k(x)~.
	\label{eq:split2}
\end{equation}
In the Equivalent Kernel limit, one sees that the action \eqref{eq:featureS} splits cleanly into greater and lesser modes, $S=S_{0}[\GPlow]+S_{0}[\GPhigh]$. At finite $\eta,\sigma^2$ however, the replicated action \eqref{eq:repS} will contain interaction terms collectively denoted $S_\mathrm{int}[\GPlow,\GPhigh]$. Explicitly, making the decomposition \eqref{eq:split1}, \eqref{eq:split2} in the replicated action \eqref{eq:repS}, we have
\begin{equation}
\begin{aligned}
    S =\,&\frac{1}{2} \sum_{m=1}^M \left(\sum_{k=1}^\kappa \frac{1}{\lambda_{k}} |f_{mk}|^2 + \sum_{k=\kappa+1}^\infty \frac{1}{\lambda_{k}} |f_{mk}|^2 \right)\\
    &-\eta\int\!\mathrm{d}\mu_x\,\exp\!\Bigg( -\frac{1}{2 \sigma^2} \sum_{m=1}^M\left[\left(\GPlow-y_<\right) +\left(\GPhigh-y_>\right)\right]^2\Bigg)~,
\end{aligned}
\label{eq:decomp2}
\end{equation}
where the first line defines the quadratic parts $S_{\mathrm{0}}[\GPlow]$ and $S_{\mathrm{0}}[\GPhigh]$, and the second line defines the interaction part $S_\mathrm{int}$. In general then, the exact partition function \eqref{eq:repZ} takes the form
\begin{equation} 
\begin{aligned}
    \langle Z^M \rangle_{\eta}&= e^{-\eta}\int\!\prod_m \mathcal{D}\GPlow \,e^{-S_0[\GPlow]} \int\prod_m\mathcal{D}\GPhigh\,e^{-S_0[\GPhigh]-S_\mathrm{int}[\GPlow,\GPhigh]} \\
    &=e^{-\eta}\int\prod_m \mathcal{D}\GPlow\,e^{-S_\mathrm{eff}[\GPlow]}~,
\end{aligned}
\label{eq:RG101}
\end{equation}
where we have defined the \emph{effective action} 
\begin{equation} 
\begin{aligned}
    S_\mathrm{eff}[\GPlow]\coloneqq-\ln\Bigg(e^{-S_0[\GPlow]} \int\prod_m\mathcal{D}\GPhigh e^{-S_0[\GPhigh]-S_\mathrm{int}[\GPlow,\GPhigh]}\Bigg)~,~ 
    \end{aligned}
 \label{eq:Seff}
\end{equation}
with $\mathcal{D}\GPlow\coloneqq\prod_{k\leq\kappa}\mathrm{d}f_{mk}$, and similarly for $\mathcal{D}\GPhigh$. In field-theoretical language, the effective action corresponds to the low-energy%
\footnote{To avoid confusion, note that the language ``high'' and ``low energy'' stems from the field-theoretic practice of working in frequency space; here, the kernel eigenmodes are akin to wavelengths, i.e., ``high-energy'' modes mean modes with low eigenvalues. \label{ft:freq}}%
effective action that describes the system up to the scale set by the cutoff,  $\kappa$.
From a Bayesian perspective, this amounts to marginalizing over the hidden degrees of freedom that are inaccessible to the observer. 

Our objective is now to perform the integral over the higher modes in \eqref{eq:Seff}. However, the presence of the interaction term complicates this, so we will first recast this term in a way which makes the interaction between the lesser and greater GP modes more tractable. To this end, we observe that the feature modes, $\phi_k(x)$, are functions over the data, which in turn is drawn from the data measure, $\mathrm{d}\mu_x=p_\mathrm{data}(x)\mathrm{d}x$. It will prove more convenient to instead treat these as fluctuating random variables over the data measure. To do this, we insert the identity in the following form into the integral over the data measure $\mathrm{d}\mu_x$ in the interaction term:
\begin{equation}
\begin{aligned}
1 &=\int\!\prod_{k=1}^\infty\mathrm{d}\varphi_k\,\delta\left(\varphi_k-\phi_k(x)\right)~,
\end{aligned}
\end{equation}
where $\varphi_k$ is a random variable over data space. The interaction term then becomes
\begin{equation}
\begin{aligned}
    S_\mathrm{int} &=-\eta\int\!\mathrm{d}\mu_x\int\!\prod_{k=1}^\infty\mathrm{d}\varphi_k \, \delta\left(\varphi_k-\phi_k(x)\right)\exp\!\left\{-\frac{1}{2 \sigma^2}\sum_{m=1}^M\left[\sum_{k=1}^\infty\left(f_{mk}-y_k\right)\phi_k(x)\right]^2\right\}\\
    &=-\eta\int\! \mathcal{D}\varphi\,P[\varphi]\,\exp\!\left\{-\frac{1}{2 \sigma^2}\sum_{m=1}^M\left[\sum_{k=1}^\infty\left(f_{mk}-y_k\right)\varphi_k\right]^2\right\}~,
\end{aligned}
\label{eq:interact2}
\end{equation}
where in the second line we implicitly defined the joint distribution over all $\varphi_k$'s as well as the new product measure, respectively,
\begin{equation}
\begin{aligned}
P[\varphi]\coloneqq& P[\varphi_1, \varphi_2, ...] = \int\!\mathrm{d}\mu_x\,
\prod^{
\infty}_{k=1}\delta\left(\varphi_k-\phi_k(x)\right)~, \qquad\label{eq:trick}\\
\mathcal{D}\varphi \coloneqq&\prod_{k=1}^\infty\mathrm{d}\varphi_k~.
\end{aligned}
\end{equation}
This removes the explicit dependence on the data, so that the new feature modes $\varphi_k$ (not to be confused with $\phi_k(x)$) are considered as random variables over feature space. Note that the second line of \eqref{eq:interact2} is now written entirely in terms of these variables. Intuitively, \eqref{eq:trick} integrates over the data measure $\mathrm{d}\mu_x$ in such a way as to ensure that the modes $\varphi_k$ correctly preserve the original distribution over feature space. 
This will enable us to capitalize on cases where $P[\varphi]$ is (approximately) Gaussian.

The features $\varphi$ can be divided into two sets $\varphi_>$ and $\varphi_<$ above and below the cutoff $\kappa$, respectively. Therefore, the joint distribution $P[\varphi]$ can be divided into probability distributions over these two sets via the conditional relation,
\begin{align} \label{eqn:varphilessgreat}
    P[\varphi]=P[\varphi_<,\varphi_>] = P[\varphi_<] P[\varphi_> | \varphi_<]~.
\end{align}

Analogous to \eqref{eq:split1}, we then define the following two variables
\begin{equation}
	\GPlm\coloneqq \sum_{k\leq\kappa}\left(f_{mk}-y_k\right) \varphi_k~,
	\quad
	\GPhm\coloneqq \sum_{k >\kappa}\left(f_{mk}-y_k\right) \varphi_{k}~.
	\label{eq:split3}
\end{equation}
For a fixed set of GP modes (i.e., within the integrand of the partition function) these can be understood as random variables over feature space. Accordingly, unlike \eqref{eq:split1}, these do not depend explicitly on the data, $x$. Rather than define separate variables as in \eqref{eq:split2}, we have also defined these to include the targets, $y_k$, since these appear together, cf. above. It will also be notationally convenient to collect these into $M$-dimensional random vectors $\GPh$, $\GPl$, so that the $m^\mathrm{th}$ component of $\GPh$ is $\GPhm$, i.e., $[\GPh]_m=\GPhm$; similarly for $\GPl$. Substituting these into \eqref{eq:interact2} and collecting terms, we have
\begin{equation}
\begin{aligned}
    S_\mathrm{int} &= -\eta\int\!\mathcal{D}\varphi\,P[\varphi]\,\exp\!\left\{-\frac{1}{2 \sigma^2}\sum_{m=1}^M\left(\GPlm+\GPhm\right)^2\right\}\\
               &=-\eta\int\!\mathcal{D}\varphi\,P[\varphi_<,\varphi_>]\,\exp\!\Bigg\{-\frac{1}{2 \sigma^2}\big(\GPl^\top\GPl+\GPh^\top\GPh+2\GPl^\top\GPh\big)\Bigg\}~\\
               &=-\eta\int\!\mathcal{D}\varphi_< \,P[\varphi_<] \exp\!\left\{-\frac{1}{2 \sigma^2}\GPl^\top\GPl\right\} \int\!\mathcal{D}\varphi_> P[\varphi_>|\varphi_<]\exp\!\left\{-\frac{1}{2 \sigma^2}\left(\GPh^\top\GPh+2\GPl^\top\GPh\right)\right\}~,
\end{aligned}
\label{eq:newsplit}
\end{equation}
where $\mathcal{D}\varphi_<\coloneqq\prod_{k\leq\kappa}\mathrm{d}\varphi_k$, and similarly for $\mathcal{D}\varphi_>$ (note that this is merely a shorthand notation, and does \emph{not} indicate a path integral). 

We can now proceed via a two-stage process, wherein we first integrate out the greater feature modes with the GP modes held fixed, and then integrate out the greater GP modes ($f_{mk}$ with $k > \kappa$). To this end, we approximate $P[\varphi]$ as a multivariate Gaussian distribution. We note in passing that a weaker requirement, namely $\GPh$ is Gaussian under the $p_\mathrm{data}(x)$ data measure, also suffices; cf. \cite{canatar}. At a technical level, this assumption allows us to perform the spatial integration in $S_\mathrm{int}$ and recast the action in terms of $f_{mk}$ alone.

Recalling that the greater and lesser feature modes are by definition linearly uncorrelated under the data measure, the multivariate Gaussian approximation implies that they are also statistically independent. Consequently, $P[\varphi_>|\varphi_<]=P[\varphi_>]={\cal N}[0,\mathbb{1};\varphi_>]$\footnote{For brevity, here we assumed that $\int\mathrm{d}\mu_x \phi_{k > \kappa}(x) = 0$, which holds in practice for many kernels. Due to the orthogonality of feature modes (cf. \eqref{eq:orthophi}), this also holds whenever the constant function is spanned by the lesser feature modes ($\phi_{k \leq \kappa}(x)$).}. The above integral over $\varphi_>$ now becomes Gaussian, and may be readily evaluated by expressing it in the form
\begin{equation}
\begin{aligned}
    \int\!&\mathcal{D}\varphi_> P[\varphi_>|\varphi_<]e^{-\frac{1}{2 \sigma^2}\left(\GPh^\top\GPh+2\GPl^\top\GPh\right)} \\
    &= \frac{1}{(2\pi)^{N_>/2}} \int\!\mathcal{D}\varphi_>e^{-\sum_{k > \kappa}\frac{\varphi^2_k}{2}}e^{-I_0[\GPh]-I_\mathrm{int}[\GPl,\GPh]}\\
  &=e^{-\frac{1}{2}\mathrm{tr}\ln\left(\mathbb{1}_{N_<\times N_<}+F^\top F/\sigma^2\right)}\exp\Bigg\{\frac{1}{2\sigma^4}\GPl^\top F\Big(\mathbb{1}_{N_<\times N_<}+F^\top F/\sigma^2\Big)^{-1}F^\top\GPl\Bigg\}~,
\end{aligned}
\label{eq:newGauss}
\end{equation}
where $F^\top F$ is an $N_>\!\times\!N_>$ matrix over greater GP modes with internal (replica) indices contracted, such that
\begin{equation}
\begin{aligned}
    I_0[\GPh]+I_\mathrm{int}[\GPl,\GPh]
    \coloneqq&\,\frac{1}{2 \sigma^2}\sum_{m;kk' > \kappa}(f_{mk}-y_k)(f_{mk'}-y_{k'}) \varphi_k \varphi_{k'}\\
    &+\frac{1}{\sigma^2}\sum_{m;k \leq \kappa,k'>\kappa}(f_{mk}-y_k)(f_{mk'}-y_{k'}) \varphi_k \varphi_{k'}\\
    \eqqcolon&\,\frac{1}{2 \sigma^2} \left(\varphi_>^\top F^\top F \varphi_> + 2 \GPl^\top F \varphi_>\right)~.
\end{aligned}\label{eq:FF}
\end{equation}
We can then express this result in terms of a covariance matrix of replica indices by Taylor expanding the logarithm in the coefficient, using the cyclicity of the trace, and re-summing the series to obtain
\begin{equation}
\begin{aligned}
   \frac{1}{2} \mathrm{tr}\ln\left(\mathbb{1}_{N_>\times N_>}+F^\top F/\sigma^2\right)
    &=\frac{1}{2}\mathrm{tr}\left(\frac{FF^T}{2\sigma^2}-\frac{(FF^\top)^2}{4\sigma^2}+\ldots \right)\\
   &=\frac{1}{2}\mathrm{tr}\ln\left(\mathbb{1}_{M\times M}+FF^\top/\sigma^2\right)\\
   &=\frac{1}{2}\mathrm{tr}\ln\left(\mathbb{1}_{M\times M}+C/\sigma^2\right)~,
\end{aligned}
\end{equation}
where $C\coloneqq FF^\top$ is the desired $M\!\times\!M$ covariance matrix with elements
\begin{equation}
	[C]_{mn}=(f_{mk}-y_k)(f_{nk}-y_k)~,
	\label{eq:covmat}
\end{equation}
cf. the definition of $F^\top\!F$ in \eqref{eq:FF}. The argument of the exponential can similarly be written in terms of this matrix by applying the Woodbury matrix identity, whereupon
\begin{equation}
		F\left(\mathbb{1}_{N_<\times N_<}+F^\top F/\sigma^2\right)^{-1}\!F^\top
		=\left(C^{-1}+\sigma^{-2}\right)^{-1}~,
		\label{eq:WoodburyFF}
\end{equation}
see appendix \ref{sec:woodbury}. Putting everything together, the full interaction term \eqref{eq:newsplit} thus becomes
\begin{equation}
\begin{aligned}
    S_\mathrm{int}=&-\eta\int\!\mathcal{D}\GPl P[\GPl]e^{-\frac{1}{2}\mathrm{tr}\ln\left(C/\sigma^{2}+\mathbb{1}\right)}\exp\left\{\frac{1}{2\sigma^2}\GPl^\top\left[\frac{1}{\sigma^{2}}\left(\sigma^{-2}+C^{-1}\right)^{-1}-\mathbb{1}\right]\GPl\right\}\\
      =&-\eta \,\,e^{-\frac{1}{2}\mathrm{tr}\ln\left(C/\sigma^{2}+\mathbb{1} \right)}\!\int\!\mathrm{d}\mu_x\,e^{-\frac{1}{2}\left(\GPlvec(x)-\bm y_<(x)\right)^\top\left(\sigma^{2}+C\right)^{-1}\left(\GPlvec(x)-\bm y_<(x)\right)}~,
\end{aligned}
\label{eq:newresult}
\end{equation}
where on the second line we have reversed our measure changes above\footnote{Explicitly, $\int\!\mathcal{D}\GPl P[\GPl]=\int\!\prod_{k\leq\kappa}\mathrm{d}\varphi_k\,P[\varphi]=\int\!\prod_{k\leq\kappa}\mathrm{d}\varphi_k\int\!\mathrm{d}\mu_x\,\delta\left(\varphi_k-\phi_k(x)\right)=\int\!\mathrm{d}\mu_x$.}, and $\GPlvec(x)$ and $\bm y_<(x)$ are $M$-dimensional vectors with elements $\GPlow(x)$ and $y_<(x)$, respectively, cf. \eqref{eq:split1} and \eqref{eq:split2}. Notably, in this expression all the higher feature modes $\phi_{k>\kappa}$ have been removed, and all higher GP modes $\GPhigh$ enter only through $C$, cf. \eqref{eq:covmat}. We note in passing that replacing $C$ with its self-consistent mean-value leads to the same average predictor as in \cite{canatar}. 

Next, we recall that, as the replica limit ($M \to 0$) is taken, our partition function should provide us with the average and variance of $f_{mk}$ as obtained by GP regression on fixed data, averaged over all datasets. Notably, the average and variance are respectively first and zeroth order in $y$ per dataset and hence must remain so after this averaging. However, in contrast to this expected result, the interaction term in our setting involves arbitrarily high powers of $y$. We argue in appendix \ref{App:NoY} that the expected linear behavior in $y$ is restored in the interaction term in the replica limit. An easy way to see this is would be to start with the interaction term having a scaled $\epsilon \,y_>$ for an infinitesimal parameter $\epsilon$, followed by divisions by the same. Any term containing second or higher order expressions in $\epsilon \, y_>$ would vanish, leaving aside linear terms in $y_>$. Following these arguments, when calculating the average predictor (which is linear in $y_<$), we may set $y_>=0$ from the start, which we do henceforth. One can then show that the contribution $\mathrm{tr}\ln\left(C/\sigma^{2}+\mathbb{1}_{M\times M}\right)$ vanishes in the replica limit. We therefore proceed with the following interaction term:
\begin{widetext}
\begin{equation}
S_\mathrm{int}=-\eta\!\int\!\mathrm{d}\mu_x\,\exp \left\{-\frac{1}{2}\left(\GPlvec(x)-\bm y_<(x)\right)^\top\left(\sigma^{2}+C_{y_>=0}\right)^{-1}\left(\GPlvec(x)-\bm y_<(x)\right)\right\}~.
\label{eq:newresult2}
\end{equation}
\end{widetext}
As a mid-derivation summary, we have now integrated out the greater modes of the {\it input measure} leading to the above rewriting of the interaction term.  However, we still need to integrate out the greater GP modes (i.e., $f_{mk}$ with $k$ above the cutoff) contained in the replica space matrix $(\sigma^2 + C_{y_>=0})^{-1}$. Despite its seemingly complex appearance, the Wilsonian RG procedure allows us to carry out this integration perturbatively as follows: we integrate out an infinitesimal momentum shell $\kappa-\delta\kappa\leq k<\kappa$, with the cutoff $\kappa$ sufficiently large such that it renders the sum of marginalized GP kernel eigenvalues $\lambda_k$ (cf. \eqref{eq:eigensys}) much smaller than the variance scale of the observation noise $\sigma^2$. That is, under the free theory for the higher modes, the expectation value of the GP covariance matrix \eqref{eq:covmat} is\footnote{To see this, observe from the first line of \eqref{eq:decomp2} that $\lambda_k$ is the variance of the Gaussian higher modes in $S_{0>}=\tfrac{1}{2}\sum_{k>\kappa}\lambda_k^{-1}f_{mk}^2$,viewed as a probability distribution in replica space. Since $[C_{y_>=0}]_{nm}=\sum_{k > \kappa} f_{mk}f_{nk}$, the expectation value $\left<C_{y_>=0}\right>_{S_0}$ is the second cumulant of this distribution.}
\begin{equation}
	\left<C_{y_>=0}\right>_{S_{0>}}=\mathbb{1}_{M\times M}\sum_{k>\kappa}\lambda_k~,
\end{equation}
where $\mathbb{1}_{nm}=\delta_{nm}$. We then set $\delta k$ such that
\begin{equation} \label{eq:deltac}
	\sum_{k=\kappa-\delta\kappa}^\kappa\lambda_k\eqqcolon\delta c\ll\sigma^2~.
\end{equation}
This ensures that at each step of the RG, $[C_{y_>=0}]_{nm}=\sum_{k} f_{mk}f_{nk}$ has a typical scale of $\delta c$. Since $\delta c/\sigma^2\ll1$, we may expand the interaction term, keeping only the leading-order contributions in $C_{y_>=0}$. One can then integrate out the greater GP modes to leading order in $C_{y_>=0}$, which amounts to replacing $C_{y_>=0}$ in \eqref{eq:newresult2} by its average according to the free theory of the greater modes, namely $C_{y_>=0} \rightarrow \delta c\, \mathbb{1}_{M \times M}$.
We thus find a renormalized ridge parameter $\sigma'$, given by
\begin{equation}
	{\sigma'}^{2} = \sigma^2 + \delta c ~.
\end{equation}
We may now repeat this procedure iteratively, as long as $\delta c$ is smaller than the effective ridge parameter. Denoting by $c$ the sum over all $\delta c$'s integrated on all shells and $\sigma^2_c$ the renormalized ridge parameter at the variance scale defined by $c$, we obtain a very simple RG flow 
\begin{equation}
	\sigma_c^2=\sigma^2+c~.
\end{equation}
In the language of RG, $\sigma^2$ is the bare ridge parameter at which the first infinitesimal shell was defined, and $\sigma_c^2$ is the renormalized value. Of course, this procedure breaks down when the smallest $\lambda$ which remains become comparable to $\sigma_c^2$. However, long before this point, we will reach the limit $\lambda = \sigma_c^2/\eta$, which marks the threshold for learnability (cf. the discussion at the end of sec. \ref{sec:background}). Thus, for a generic target function, a natural point to stop the momentum shell RG flow would be at the first learnable mode.

It is interesting to compare this result with other works which use the feature Gaussianity assumption, such as \cite{canatar}. There it was found that the dataset average predictor coincides with the Equivalent Kernel approximation with an effective ridge parameter, $\kappa$. This is similar to our result in that their $\kappa$ is the original ridge parameter plus all the eigenfunctions which are unlearnable. There are two important conceptual differences however: in that derivation, there is no notion of RG, nor of a Wilsonian cut-off or RG flows. The same largely holds for their concurrent work \cite{atanasov2024scaling}. Furthermore, the criteria for absorbing unlearnable modes is not sharp as in our case, but is instead controlled by the ratio of $\kappa$ over the number of datapoints and the eigenfunctions.

In the next section, we will show that the Gaussian $P[\varphi]$ assumption leads to reasonable predictions for two real-world datasets (MNIST and CIFAR10). In section~\ref{sec:beyond}, we will study how the picture changes when $P[\varphi]$ is allowed to be slightly non-Gaussian. As an illustrative empirical example, we then show how this non-Gaussian treatment can capture a qualitative effect which would otherwise be missed (section~\ref{sec:toy}).

\section{\label{sec:scalingLaws} Neural Scaling Laws}
A variety of DNNs exhibit robust semi power-law dependence on data, parameters, and compute budget \cite{kaplan2020scaling} making their potential improvement with scale highly predictable. In the context of GPs, this behavior has been explained in \cite{bahri2021explaining} and emanates from the remarkable fact that many real-world datasets exhibit a power-law distribution of kernel eigenvalues; specifically, the fact that the bulk of eigenvalues scale as $\lambda_k \propto k^{-(1+\alpha)}$, with $\alpha>0$ being a dataset-dependent exponent.   
Within the EK treatment, we can analyze the decay of the MSE loss, by noting that the $50\%$ learnability threshold occurs when the average number of data points is $\eta_T=\sigma^2/\lambda_{k_T}$ and hence at $k_T=(\eta_T/\sigma^2)^{1/(1+\alpha)}$. 
Assuming the target function matches the GP namely $y^2_k \propto k^{-(1+\alpha)}$, we can estimate the loss via $\sum_{k > k_T}^{\infty} y^2_k \approx \int_{k_T}^{\infty} dk k^{-(1+\alpha)} \propto \eta_T^{-\alpha/(1+\alpha)}$. Therefore, we find a so-called data exponent of $\alpha_D=\alpha/(1+\alpha)$. 

Next, let us provide a heuristic of how our RG setup changes this scaling for $\sigma^2=0$. After integrating out all modes down to $k_{\Lambda}$ cutoff, we obtain $\sigma^2_{\text{eff}}=\sum_{k_{\Lambda}}^{\infty} k^{-(1+\alpha)} \approx \alpha^{-1} k_{
\Lambda}^{-\alpha}$. The RG flow remains tractable in the sense of maintaining some sufficiently small $\delta c$, as we have $\sigma^2_{\text{eff}}(k_{\Lambda}) \propto k_{\Lambda}^{-\alpha} \gg \lambda_{k_{\Lambda}}$. At this point, we undertake a self-consistent assumption that the growth of $\sigma_{\text{eff}}^2$ puts us sufficiently close to the EK limit so that we can evaluate the threshold for learnable modes via $\eta_T=\sigma_{\text{eff}}^2(k_T)/\lambda_{k_T}\approx \alpha^{-1} k_T^{-\alpha}/k_T^{-(1+\alpha)}=\alpha^{-1}k_T$. This leads to an unexpectedly reasonable result which shows that the number of learnable modes scales linearly with the dataset size. Evaluating the MSE loss by resumming all unlearnable modes of $y_k$, we now obtain $\sum_{k > k_T}^{\infty} y^2_k \approx \int_{k_T}^{\infty} dk k^{-(1+\alpha)} \propto k_T^{-\alpha} \propto \eta_T^{-\alpha}$, which is the data exponent $\alpha_D$ found in \cite{bahri2021explaining} using asymptotic properties of the hypergeometric function obtained from the explicit solution of the self-consistent equations of \cite{canatar}. 

After this, we verify the closeness of the system to the EK limit. We highlight here that the EK approximation comes from the Taylor expansion of $\int d\mu_x e^{-(f(x)-y(x))^2/2\sigma^2}$. Therefore, we need to compare $\sigma_{\text{eff}}^2$ with the mean MSE loss that we have obtained. The relation $\sigma_{\text{eff}}^2=\alpha^{-1} k_T^{-\alpha}$ places $\sigma^2_{\text{eff}}$ at the same scale as the loss thus placing the system at the threshold of the EK approximation, where it is expected to provide a correct order of magnitude estimate. The task of further developing this, through an analysis of the RG-relevancy of the higher order terms obtained from the above Taylor expansion, is left for future work.

\subsection{\label{sec:scalingLaws_empirical} Empirical Demonstration}
In this section we further corroborate the claim in 
section~\ref{sec:scalingLaws} that our RG method is capable of predicting the power-law scaling behavior of the loss as a function of dataset size, which has been observed in real-world datasets~\cite{bahri2021explaining}. We consider one of the regression tasks from Ref.~\cite{canatar} on two such datasets, MNIST~\cite{MNIST} and CIFAR10~\cite{CIFAR10}. In these experiments, we compare the empirical scaling of the MSE loss to the predictions from our RG theory\footnote{Our RG theory prediction involves two steps: (1) we find the renormalized noise, $\sigma_{eff}^2$, arising from unlearnable modes; (2) we use this renormalized noise, as opposed to the typical $\sigma^2$, in the EK approximation formula \eqref{eq:naive_EK_avpred}. Thus, by RG theory, we mean the EK prediction modified with a renormalized noise, $\sigma^2 \rightarrow \sigma_{eff}^2$.}, the naive EK approximation (in which the noise, $\sigma^2$, is not renormalized), and the current state-of-the-art Spectral Bias theory~\cite{canatar}. We find that our RG theory accurately predicts the empirical power-law scaling of the MSE loss on these real-world datasets.

\begin{figure}
    \centering
    \includegraphics[width=0.49\linewidth]{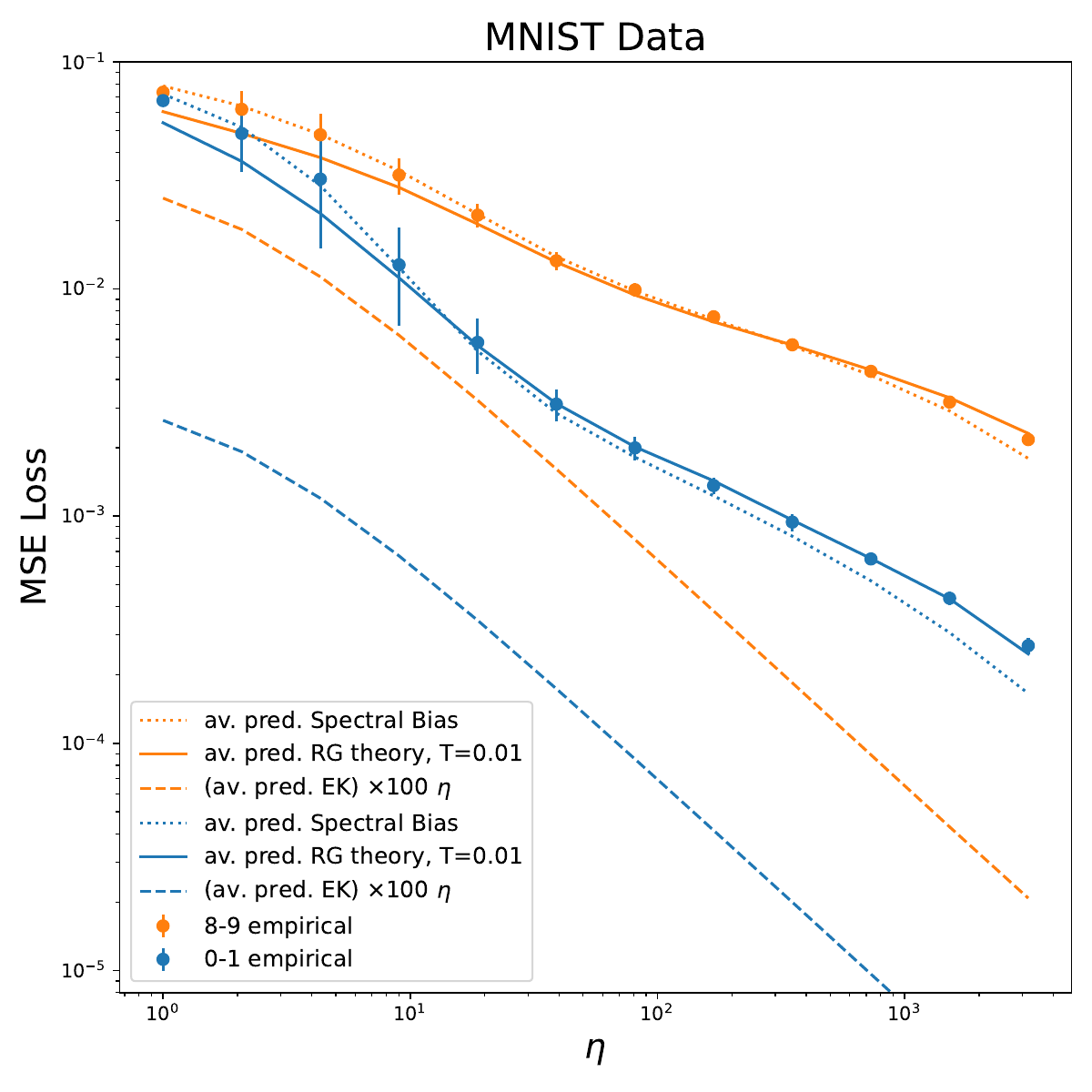}
    \includegraphics[width=0.49\linewidth]{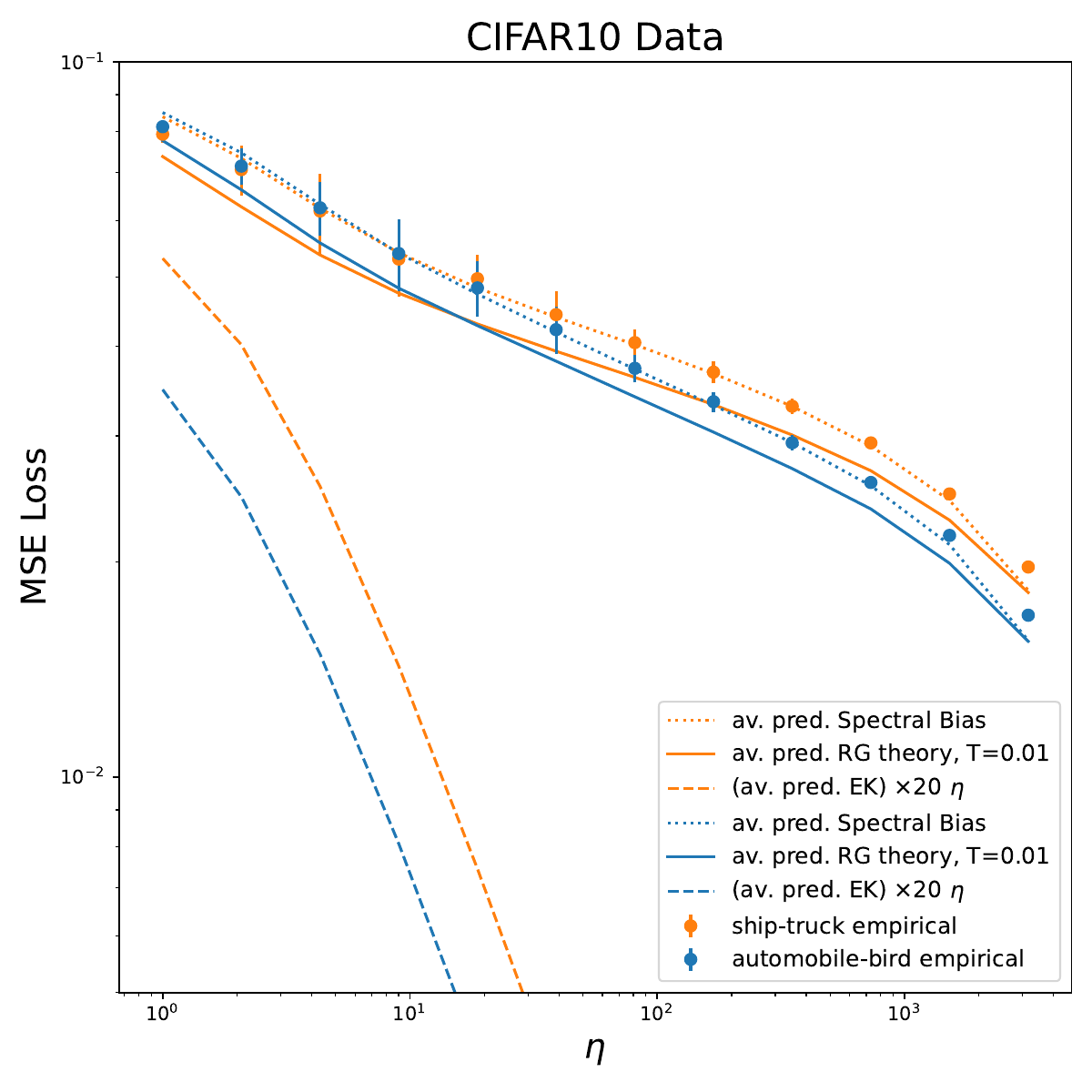}
    \caption{ \textbf{Predicting empirical MSE loss scaling on MNIST and CIFAR10 regression tasks.} For two real-world dataset examples (MNIST and CIFAR10), we show different theoretical predictions for the MSE loss as a function of the number of datapoints, $\eta$. For each dataset, we consider two different regression tasks. For MNIST we consider both `8-9' and `0-1', where $N=9,837$ and $N=10,564$, respectively. For CIFAR10 we consider both `ship-truck' and `automobile-bird', where $N=10,000$ in both cases. For all experiments, we fix $\sigma^2=10^{-8}$. We find that both the state-of-the-art Spectral Bias theory~\cite{canatar} and the RG theory introduced in this work predict the semi power-law behavior well. Whereas, the EK approximation in which there is no noise renormalization (i.e. $\sigma_{\rm eff}^2 = \sigma^{2}$) fails to accurately predict this behavior.}
    \label{fig:empirical_results_scalinglaws}
\end{figure}

MNIST and CIFAR10 are both datasets with 10 classes of image types. In what follows, we closely follow the setup of Ref.~\cite{canatar}. We consider $N$ samples with approximately half belonging to one of these 10 classes, and half to another. For example, for MNIST, we consider two separate tasks `8' vs `9' regression as well as `0' vs `1'. Thus, our measure is explicitly  $d\mu_x = p_{\rm data}(x) dx$ with $p_{\rm data}(x) = \frac{1}{N} \sum_{i=1}^{N} \delta(x - x_i)$ where every $x_i$ corresponds to e.g. an `8' or a `9' MNIST image.  We consider the Neural Tangent Kernel (NTK)~\cite{jacot2018neural} limit of a fully connected network (FCN) with ReLU activations on hidden layers; we use the \texttt{neural-tangents} package~\cite{neuraltangents2020, novak2022fast, hron2020infinite, sohl2020on, han2022fast} to obtain the kernel function, $K(x,x')$. For each task, we solve for the kernel's eigenvalue spectrum, $\lambda_k$, and corresponding eigenfunctions, $\phi_k(x)$, via~\eqref{eq:eigensys}.  
We then use the eigenfunctions to decompose the target, $y(x)$, in terms of the coefficients, $y_k$, as in~\eqref{eq:specdecomp} (i.e. $y_k = \int d\mu_x ~\phi_k(x) ~y(x)$). 
These can similarly be used to construct our theoretical average prediction, $\hat{y}(x)$,
\begin{align}
    \hat{y}(x) = \sum_{k=1}^{N} ~\phi_k(x) ~\bar{f}_k = \sum_{k=1}^{N} \left(\frac{\lambda_k}{\lambda_k + \sigma^2_{\rm eff}/\eta}\right)~\phi_k(x)  ~y_k.
\end{align}
Comparing with~\eqref{eq:naive_EK_avpred}, we can see that the key difference from the naive EK is the noise renormalization i.e. $\sigma^2 \rightarrow \sigma_{eff}^2$. We can then express the theoretical MSE loss as follows
\begin{align}\label{eq:mse_loss}
    {\rm MSE} \left(y,\hat{y}\right) = \frac{1}{N} \sum_{k=1}^{N} \lvert ~\bar{f}_k - y_k\rvert^2 
    = \frac{1}{N} \sum_{k=1}^{N} L_k^2  ~y_k^2,
\end{align}
where we have implicitly defined the \emph{learnability factor},
\begin{align}\label{eq:learnabilityfactor}
L_k := \frac{\sigma^2_{\rm eff}}{\eta \lambda_k + \sigma^2_{\rm eff}}.
\end{align}
In section~\ref{sec:scalingLaws}, we assumed that $L_k=0$ for a learnable mode ($k\leq\kappa$) and $L_k=1$ for an unlearnable mode ($k>\kappa$). Here we will offer a more careful treatment using~\eqref{eq:learnabilityfactor}. Fixing a learnability threshold, $T \in (0,1)$, we numerically find the number of learnable modes, $\kappa$, such that $L_\kappa \approx T$\footnote{In appendix~\ref{app:real_data_vary_threshold}, we explore the sensitivity to different threshold choices.}. Note that we assume $1\leq \kappa \leq N$ (i.e. that there is at least one learnable mode). Thus, $\kappa$ generally will depend both on the number of observations, $\eta$, as well as the initial noise, $\sigma^2$, which enters through $L_k$'s dependence on $\sigma_{\rm eff}^2 = \sigma^2 + \sum_{k=\kappa+1}^{N} \lambda_k$. With this critical $\kappa_T$, we then calculate our prediction for the loss according to~\eqref{eq:mse_loss}. Repeating this procedure for increasing numbers of observations, $\eta$, gives us a theoretical prediction of how the MSE loss scales as a function of dataset size. In Fig.~\ref{fig:empirical_results_scalinglaws}, we compare these theoretical predictions to empirical GP kernel regression results obtained using the scheme described in Ref.~\cite{canatar}. Indeed, we find that our RG theory accurately predicts the power-law dependence of the loss on $\eta$, and is a great improvement over the naive EK prediction (i.e.,~\eqref{eq:naive_EK_avpred}).

\begin{figure}
    \centering
    \includegraphics[width=0.49\linewidth]{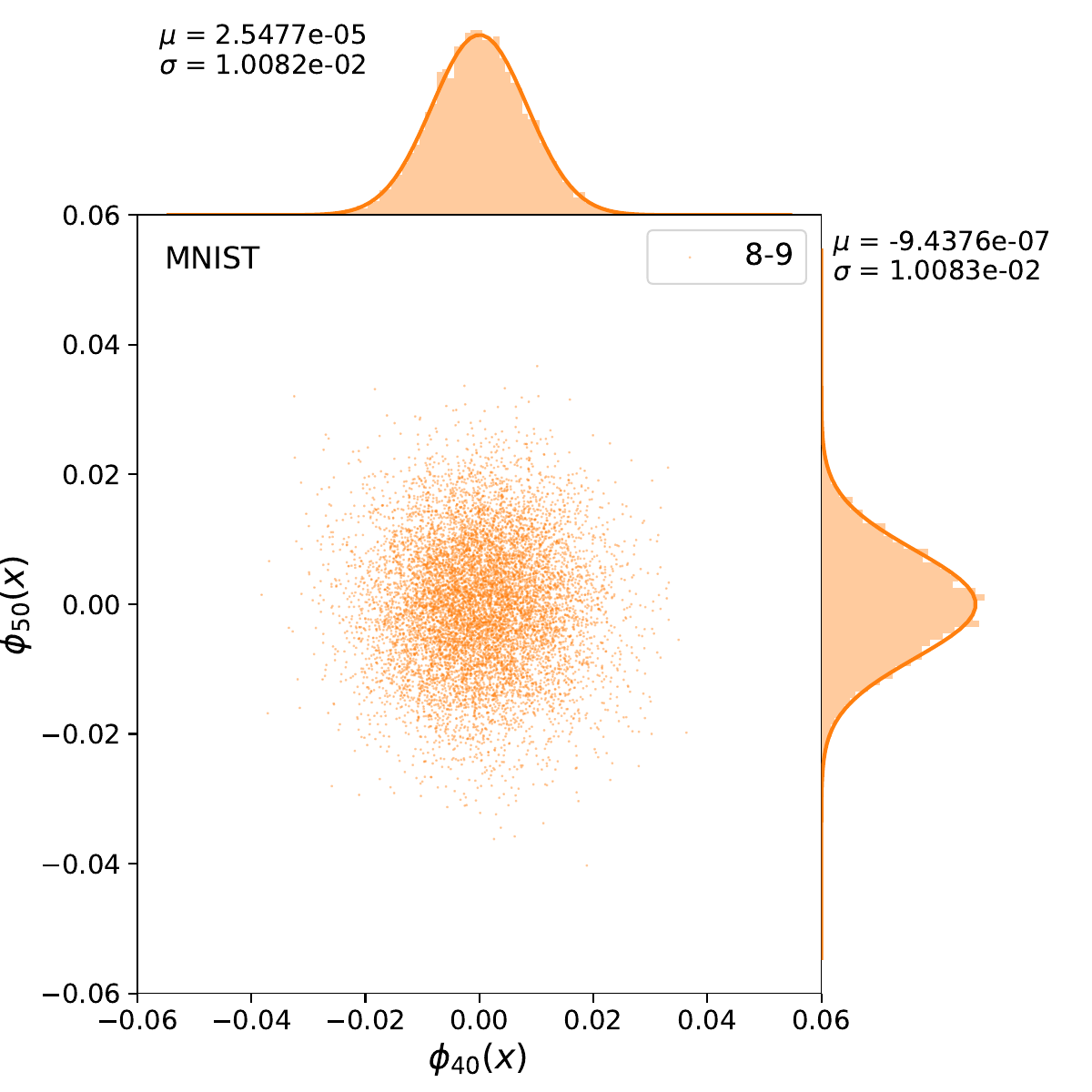}
    \includegraphics[width=0.49\linewidth]{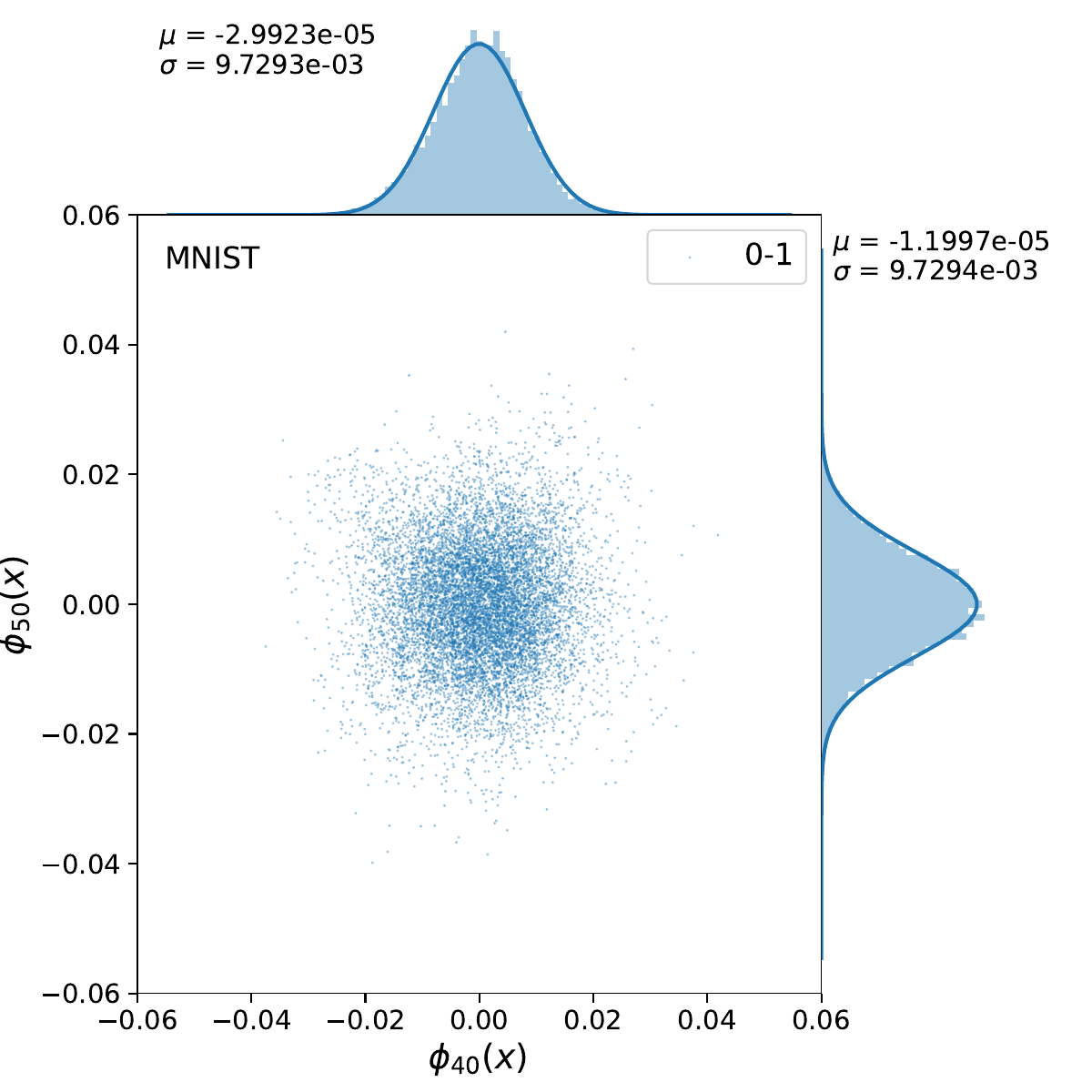}
    \includegraphics[width=0.49\linewidth]{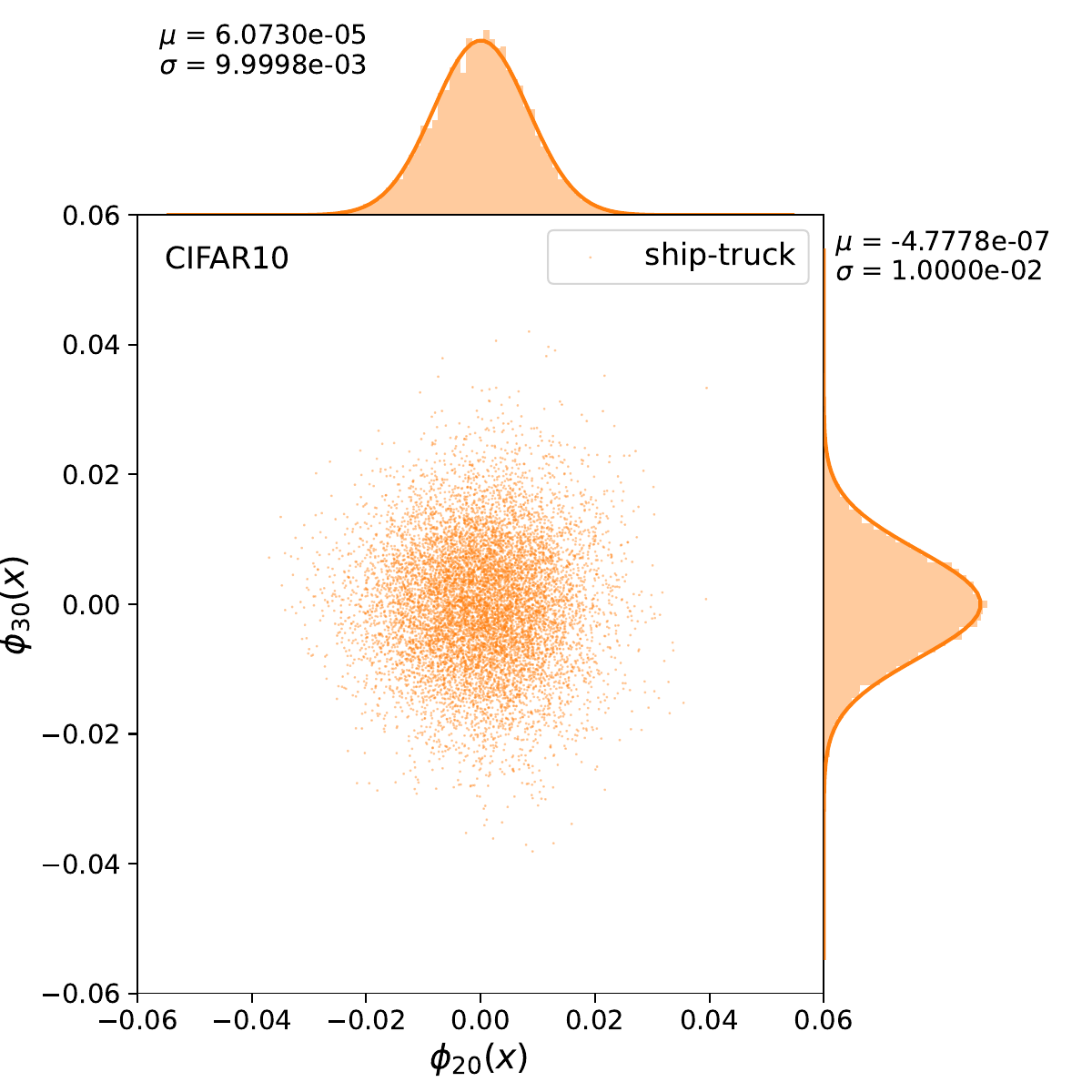}
    \includegraphics[width=0.49\linewidth]{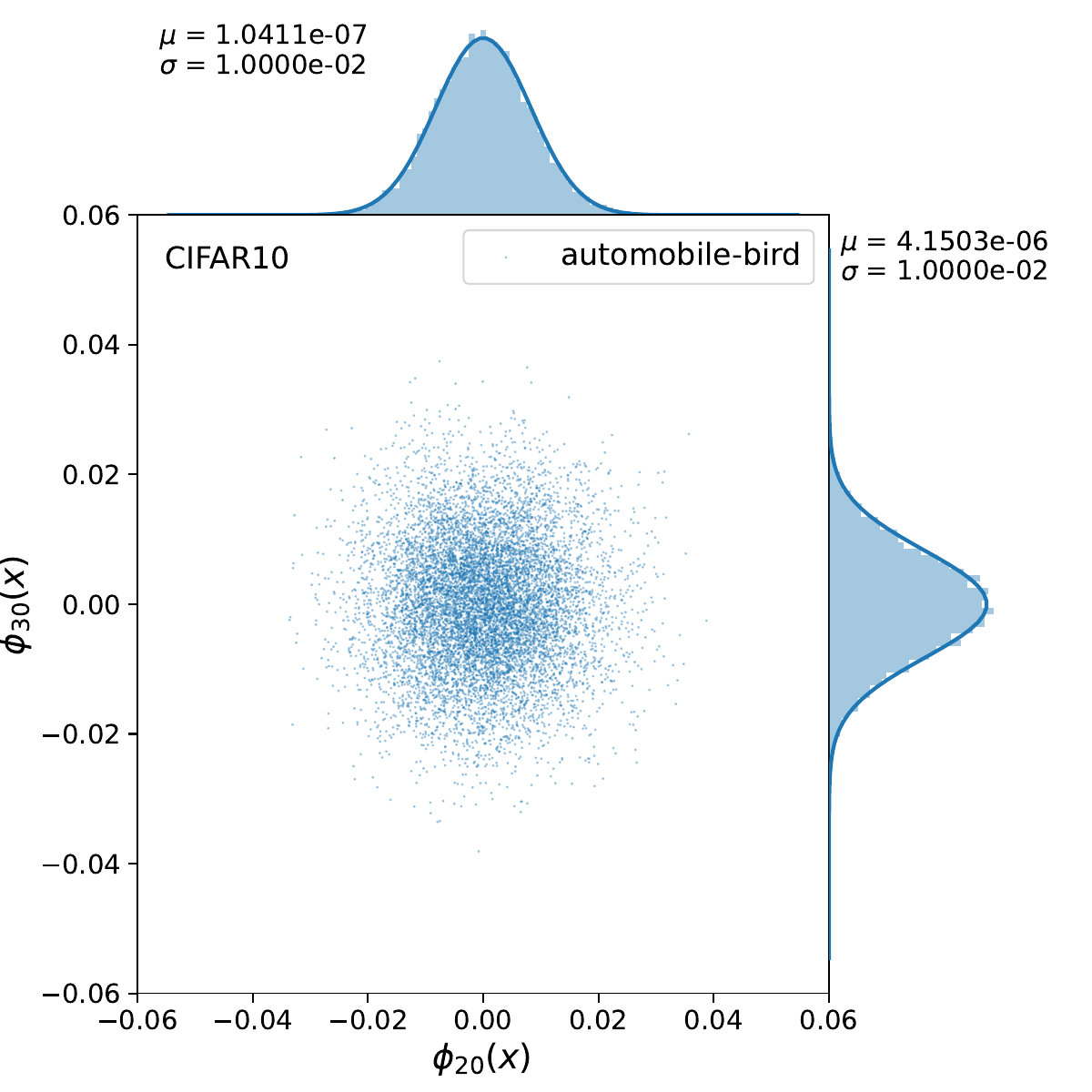}
    \caption{ \textbf{Illustrative depiction that the feature modes of real-world datasets are often Gaussian distributed.} For two real-world dataset examples (MNIST and CIFAR10), we select two feature modes and show their joint distribution along with the best Gaussian fit to each marginal. This is done for each learning task considered. Namely, `8-9' and `0-1' in the MNIST case and `ship-truck' and `automobile-bird' in the CIFAR10 case. As previously noted in~\cite{Simon2021}, we can observe the interesting fact that the feature modes of real-world datasets are often (jointly) Gaussian distributed. Appendix~\ref{app:gaussian_vs_cauchy} explores the extent to which this varies as a function of $k$.}    
    \label{fig:empirical_results_scalinglaws_gaussianfeaturesproof}
\end{figure}

\section{\label{sec:beyond} Renormalization beyond the Gaussian}

Here we extend the RG formalism to accommodate non-Gaussian effects associated with the feature modes, namely, cases where the $\varphi$ random variables (under the data measure) have substantial deviations from normality. We begin with a somewhat artificial but illustrative scenario where the greater features are non-Gaussian yet remain independent from the lesser features; that is, $P[\varphi_>|\varphi_<]=P[\varphi_>]$ with $P[\varphi_>]$ non-Gaussian. 
In this case, we show that the non-Gaussianity of $P[\varphi_>]$ has no effect, and thus the previous treatment still holds without any changes. We then proceed with allowing weak non-Gaussianity with potential dependencies between greater and lesser feature modes. Schematically, this corresponds to $P[\varphi_<,\varphi_>] \approx {\cal N}[0,\mathbb{1};\varphi_<]{\cal N}[0,\mathbb{1};\varphi_>] + \text{corrections} \Rightarrow P[\varphi_>|\varphi_<] \approx {\cal N}[0,\mathbb{1};\varphi_>] + \text{corrections}$, where the corrections are to be made precise below. This corrected form of $P[\varphi_>|\varphi_<]$ is subsequently used in~\eqref{eq:newsplit} where, alongside the noise renormalization, we obtain a spatial re-weighting of the mean squared error (MSE) loss or, equivalently, a spatial re-weighting of the ridge parameter. 

Considering the first case in which the greater and lesser features are independent, the conditional probability in \eqref{eq:newsplit} equals the marginal probability, $P[\varphi_>|\varphi_<]=P[\varphi_>]$, so that the integral over greater modes becomes
\begin{equation}
	\int\!\mathcal{D}\varphi_> P[\varphi_>]\exp\!\left\{-\frac{1}{2 \sigma^2}\left(\GPh^\top\GPh+2\GPl^\top\GPh\right)\right\}~.
	\label{eq:artificial}
\end{equation}
In accordance with our perturbative RG prescription above, we take $\GPh$ to contain a single feature mode (the rest are to be integrated out in subsequent steps), which we denote by $\varphi_q$, so that $\GPq$ is the replica space vector whose $m^\mathrm{th}$ element is $\Phi_{mq}\coloneqq[\Phi_q]_m=(f_{mq}-y_q)\varphi_q$, cf. \eqref{eq:split3}. The above integral can then be viewed as the cumulant generating function of $P[\varphi_q] \exp\{-\frac{1}{2 \sigma^2}\GPq^\top \GPq \!\}$ with $-\GPq^\top \GPl/\sigma^2$ playing the role of the cumulant function argument. Accordingly, the leading order non-Gaussian corrections will enter as $(f_<(x)-y_<(x))^4$ times the fourth cumulant of $\varphi_q$. For reasonable (e.g., sub-Gaussian) distributions\footnote{Since our approach is perturbative, we limit our study to distributions without heavy tails, cf. \cite{zavatoneveth2021exact}.}, the  (2n)$^\mathrm{th}$ cumulant scales like the variance to the $n^\mathrm{th}$ power. The higher cumulants are thus $O\left(\delta c^2\right)$ and hence negligible, and we obtain the same equations as in the Gaussian case.

We now turn to the more realistic case in which the lesser and greater feature modes are coupled, and relax the Gaussianity assumption by noting that if the data has a high effective dimension, $d$, then a generic feature -- which is a sum of many monomials in $x_i$ -- can be viewed as a sum of many weakly correlated random variables and hence well-approximated by a Gaussian. This, together with the empirical success of the Gaussian assumption \cite{canatar}, motivates us to consider corrections to Gaussianity as small $1/d$ corrections (as in \cite{tomasini2022failure}) and treat the latter perturbatively, analogous to the 1/N expansion in quantum field theory. At a technical level, we do so by keeping a leading-order Edgeworth expansion of $P[\varphi_<,\varphi_>]$. We then use this expansion to obtain the conditional distribution $P[\varphi_>|\varphi_<]$ appearing in \eqref{eq:newsplit}, and perform the integral perturbatively by keeping only the leading non-Gaussianities. Concretely, we take as our large-$d$ expansion 
\begin{equation}
	P[\varphi_<,\varphi_>] \approx \;{\cal N}[0,\mathbb{1};\varphi_<]{\cal N}[0,\mathbb{1};\varphi_>] 
\Bigg(1 + \frac{1}{4!} \sum_{\substack{k_1,k_2,\\k_3,k_4}} U_{k_1 k_2 k_3 k_4} He_4(\varphi_{k_1},\varphi_{k_2},\varphi_{k_3},\varphi_{k_4})\Bigg)~,\quad
\label{eq:Edge}
\end{equation}
where ${\cal N}[0,\mathbb{1};\varphi_*]$ denotes a standard normal distribution of the feature mode vector $\varphi_*$, $He_4$ is the fourth multivariate (probabilist's) Hermite polynomial,
\begin{equation}
\begin{aligned}
	He_4(\varphi_{k_1},\varphi_{k_2},\varphi_{k_3},\varphi_{k_4}) 
	=&\; {\cal N}[0,\mathbb{1};\varphi_<]^{-1}{\cal N}[0,\mathbb{1};\varphi_>]^{-1} 
\partial_{\varphi_{k_1}}\partial_{\varphi_{k_2}}\partial_{\varphi_{k_3}}\partial_{\varphi_{k_4}}{\cal N}[0,\mathbb{1};\varphi_<]{\cal N}[0,\mathbb{1};\varphi_>]~,\quad
\end{aligned}
\end{equation}
(see for example \cite{Naveh_2021}), and $U$ is the (fourth) cumulant/Ursell function,
\begin{equation}
	U_{k_1 k_2 k_3 k_4} \coloneqq \langle \varphi_{k_1} \varphi_{k_2} \varphi_{k_3} \varphi_{k_4} \rangle_{P[\varphi],\mathrm{connected}}~,
\end{equation}
i.e., the connected 4-pt correlation function with respect to $P[\varphi]$ \footnote{Note that this is the exact 4-pt function, which takes into account any potential RG effects in the input/$P[\varphi]$ distribution such as those considered in Ref.~\cite{Bradde_2017}.}. We emphasize that in our perturbative expansion in $1/d$, we shall keep only terms which are leading order in the fourth cumulant, again in analogy with the $1/N$ expansion, in which the connected $2k$-point correlation function is suppressed by $1/N^{k-1}$. Notably the lack of odd cumulants is consistent with our previous simplifying assumption that all features are odd functions therefore $P[\varphi]=P[-\varphi]$. In general, we think of $\varphi$ as spanning some relevant (i.e., potentially learnable) subspace of function space and do not require this subspace to be closed under the multiplication of features (e.g., $\phi_{k}(x)^2$, is an even function and thus outside this subspace). A different set of assumptions is taken in sec. (\ref{sec:toy}), leading to similar results.

As per our RG prescription above, we again take $\GPh$ to consist of a single feature mode $\varphi_q$, whose corresponding GP mode $f_{mq}$ has a variance $\delta c$ much smaller than $\sigma_c^2$. Let us examine the possible fourth cumulants, $U_{k_1,k_2,k_3,k_4}$, $U_{k_1,k_2,k_3,q}$, $U_{k_1,k_2,q,q}$, $U_{k_1,q,q,q}$, and $U_{q,q,q,q}$, where henceforth $k_i$ refers to lesser modes. Since the variance of the higher mode scales like $\delta c$, the last of these scales like $\delta c^2$, while the second to last scales like $\delta c^{3/2}$. Hence, keeping only the leading order terms in $\delta c$ and $1/d$, we arrive at the following RG-equivalent description of \eqref{eq:Edge}:
\begin{equation}
\begin{aligned}
    P[\varphi_<,\varphi_q]
     \approx&\; {\cal N}[0,\mathbb{1};\varphi_<]\left(1+\frac{1}{4!}U_{k1 k_2 k_3 k_4} He_4(\varphi_{k_1},\varphi_{k_2},\varphi_{k_3},\varphi_{k_4}) \right) \\
                   &\times {\cal N}[0,\mathbb{1};\varphi_q] \Bigg(1+ \frac{1}{3!} U_{k_1 k_2 k_3 q} He_3(\varphi_{k_1},\varphi_{k_2},\varphi_{k_3}) \varphi_q \\
    & \qquad \qquad \qquad \quad + \frac{1}{2!} U_{k_1 k_2 q q} He_2(\varphi_{k_1},\varphi_{k_2}) [\varphi_q^2 - 1] \Bigg)~, 
\end{aligned}
\label{eq:LowHighSplitHermite}
\end{equation}
where we have adopted Einstein summation notion, and used relations between multivariate Hermite polynomials implying, e.g., that $He_4(\varphi_{k_1},\varphi_{k_2},\varphi_{k_3},\varphi_q)=He_3(\varphi_{k_1},\varphi_{k_2},\varphi_{k_3})He_1(\varphi_q)$. The combinatorial factors arise from the various ways of fixing $q$. Concretely, the 2nd and 3rd Hermite polynomials for the lesser modes appearing in this expression are
\begin{equation}
\begin{aligned}
    He_2(\varphi_{k_1},\varphi_{k_2}) &= \varphi_{k_1}\varphi_{k_2}-\delta_{k_1,k_2}~,\\
    He_3(\varphi_{k_1},\varphi_{k_2},\varphi_{k_3}) &= \varphi_{k_1}\varphi_{k_2}\varphi_{k_3}-\delta_{k_1,k_2}\varphi_{k_3}-\delta_{k_2,k_3}\varphi_{k_1}-\delta_{k_1,k_3}\varphi_{k_2}~,
\end{aligned}
\end{equation}
while 
\begin{equation}
	He_1(\varphi_q)=\varphi_q~,
	\qquad
	He_2(\varphi_q^2)=\varphi_q^2-1~.
	\label{eq:qHermite}
\end{equation}
Finally we note that, since we perform the Edgeworth expansion about the standard normal $\mathcal{N}[0,\mathbb{1};\varphi_q]$, both the Hermite polynomials of the higher modes \eqref{eq:qHermite} have zero mean under this distribution. Consequently, the top line in \eqref{eq:LowHighSplitHermite} is the marginal distribution of the lesser feature modes, and therefore
\begin{equation}
    P[\varphi_q|\varphi_<] \equiv \frac{P[\varphi_q,\varphi_<]}{P[\varphi_<]}\approx {\cal N}[0,\mathbb{1};\varphi_q] \left[1+ A\varphi_q + B\left(\varphi_q^2 - 1\right)\right],
\label{eq:conEdge}
\end{equation}
where we have introduced the shorthand notation
\begin{equation}
\begin{aligned}
	A &\coloneqq \frac{1}{3!} U_{k_1 k_2 k_3 q} He_3(\varphi_{k_1},\varphi_{k_2},\varphi_{k_3}), \\
	B &\coloneqq \frac{1}{2!} U_{k_1 k_2 q q} He_2(\varphi_{k_1},\varphi_{k_2}),
\end{aligned}
\end{equation}
where the first non-trivial term constitutes a shift of the conditional mean, and the second is a shift of the conditional variance of the greater mode. 

Now, proceeding from the integral in \eqref{eq:newsplit}, we will substitute the expansion \eqref{eq:conEdge} in place of $P[\varphi_>|\varphi_<]$. First however, we note that as per our earlier assumption on the relative variance of $\GPh$, i.e., $\delta c\ll\sigma^2$, we may neglect the $\GPh^\top\GPh$ term in the exponential, so that\footnote{Alternatively, one can retain this term throughout the subsequent calculation and show that this amounts to neglecting $O\!\left(\delta c^2/\sigma^4\right)$ terms.} 
\begin{equation}
\int\!\mathcal{D}\varphi_>P[\varphi_>|\varphi_<]\,e^{-\frac{2 \GPl^\top \GPh +\GPh^\top \GPh }{2 \sigma^2}}\approx \int\!\mathcal{D}\varphi_>P[\varphi_>|\varphi_<]\,e^{-\frac{\GPl^\top \GPh} {\sigma^2}}.
\label{eq:potato}
\end{equation}
Additionally, since we take $\GPh$ to consist of a single feature mode, we keep only a single term in the decomposition \eqref{eq:split3}, so that at each step of the RG, the integral \eqref{eq:potato} becomes 
\begin{equation}
\int\!\mathcal{D}\varphi_>P[\varphi_>|\varphi_<]\,e^{-\frac{\GPl^\top \GPh} {\sigma^2}} \;\;\overset{\textrm{RG shell}}{\longrightarrow}\;\;
\int\!\mathcal{D}\varphi_q P[\varphi_q|\varphi_<] e^{-\frac{\GPl^\top\GPq} { \sigma^2}}~.
\label{eq:usethisone}
\end{equation}

Hence, substituting the large-$d$ expansion \eqref{eq:conEdge} in place of the conditional distribution and re-exponentiating after performing the Gaussian integral, we obtain
\begin{equation}
    \begin{aligned}
    \int\!\mathcal{D}\varphi_q &P[\varphi_q|\varphi_<] e^{-\frac{\GPl^\top\GPq} { \sigma^2}}
    =\frac{1}{\sqrt{2\pi}}\int\!\mathcal{D}\varphi_q  e^{-\frac{\varphi^2_q}{2}- \frac{\GPl^\top\GPq}{\sigma^2}}\left[1+ A\varphi_q + B\left(\varphi_q^2 - 1\right)\right]\\
        &=\frac{1}{\sqrt{2\pi}}\int\!\mathcal{D}\varphi_q  e^{-\frac{\varphi^2_q}{2}- \frac{1}{\sigma^2}\sum_m\GPlm\GPmq} \left[1+ A\varphi_q + B\left(\varphi_q^2 - 1\right)\right]\\
        &\approx \exp\left[\frac{1+2B}{2\sigma^4}\left(\GPl^\top\fq\right)^2 -\frac{A}{\sigma^2}\GPl^\top \fq\right]~,
\end{aligned}\label{eq:shorter}
\end{equation}
where we have defined $\fq$ to be the replica-space vector whose $m^\mathrm{th}$ element is $[\fq]_m=f_{mq}$, cf. the definition of $\Phi_q$ below \eqref{eq:artificial}, and the approximation refers to neglecting terms higher-order in $1/d$. 

Our next task is to integrate out the greater GP mode ($f_{mq}$), again to leading order in the non-Gaussianity (consistent with our Edgeworth expansion above). Recalling the replicated partition function from sec. \ref{sec:Gaussian}, we have 
\begin{widetext}
\begin{align}
	\left<Z^M\right>_\eta=e^{-\eta}\int\!\mathcal{D}\bm f\,\exp \left\{-S_0[\bm f]+\eta\int\!\mathcal{D}\varphi P[\varphi]\,\exp\left[-\frac{1}{2\sigma^2}\left(\GPl^\top\GPl+2\GPl^\top\GPh+\GPh^\top \GPh\right)\right] \right\}
		\label{eq:path2},
  \end{align}
\end{widetext}
where $\mathcal{D}\bm f\coloneqq\prod_m\mathcal{D}f_m=\prod_{mk}\mathrm{d}f_{mk}$ (cf. \eqref{eq:RG101}), and $S_0$ denotes the free action given in the first line of \eqref{eq:decomp2}, which we repeat here for convenience:
\begin{equation*}
	S_0[f]=\frac{1}{2}\sum_{m=1}^M\sum_{k=1}^\infty\frac{1}{\lambda_k}|f_{mk}|^2~.
\end{equation*}
By \eqref{eq:newsplit} and the above analysis leading to \eqref{eq:shorter}, the second term in the exponential of $S_{\mathrm{int}}$ becomes
\begin{equation}
 \begin{aligned}
    S_{\mathrm{int}} =
    &-\eta\int\!\mathcal{D}\varphi P[\varphi]\,\exp\left[-\frac{1}{2\sigma^2}\left(\GPl^\top\GPl+2\GPl^\top\GPh +\GPh^\top \GPh\right)\right]\\
      &\approx-\eta\int\!\mathcal{D}\varphi_<\,P[\varphi_<]\,\exp\left[-\frac{1}{2\sigma^2}\GPl^\top\GPl+\frac{1+2B}{2\sigma^4}\left(\GPl^\top \fq\right)^2 -\frac{A}{\sigma^2}\,\GPl^\top \fq \right]~.
\end{aligned}
\label{eq:strawberry}
\end{equation}
Splitting the path integral as in \eqref{eq:RG101}, we thus have
\begin{equation} 
	\left<Z^M\right>_\eta=e^{-\eta}\int\!\mathcal{D}\bm f_<\,e^{-S_0[\bm f_<]}\int\!\mathcal{D}\fq\,e^{-S_0[\fq] - S_{\mathrm{int}}}~.
\end{equation}
Now, to compute the path integral over $\fq$, we again work perturbatively, keeping terms up to order $\fq^2/\sigma^2 \sim \delta c/\sigma^2$, and only linear order in $A$ and $B$ (as per our large-$d$ expansion). As above, this can be done by expanding the interaction terms in the action, then expanding the exponential of the partition function, performing the integral over $\fq$, and then re-exponentiating the result. To leading order in the non-Gaussianities, we find:
\begin{equation} 
    \int\!\mathcal{D}\fq\,e^{-S_0[\fq] - S_{\mathrm{int}}} \approx\exp\left\{\eta\int\!\mathcal{D}\varphi_<\,P[\varphi_<]\,e^{-\frac{\GPl^\top\GPl}{2\sigma^2}+\lambda_q(1+2B)\frac{\GPl^\top\GPl}{2\sigma^4}}\right\}~,\quad
    \label{eq:huge}
\end{equation}
where $\lambda_q\sim\delta c$, since this is the mode that is integrated-out at a single RG step. As in the Gaussian case, this results in an effective substitution of $\fq^\top\fq$ and $\fq$ with their averages under $S_0[\bm{f}_q]$, namely $\lambda_q \mathbb{1}_{M\times M}$ and $0$. As expected by the linearity of GP regression, the result is quadratic in $\GPl$. Interpreting the term in the exponent as a loss function, we see that it remains MSE in nature but becomes ``spatially weighted'' by the $1+2B[\varphi_<]$ factor. Indeed, recasting this term back into $\mathrm{d}\mu_x$ and $\phi_k(x)$ notation, 
\begin{equation}
\begin{aligned}
    1+2B[\varphi_<]&\sim1+2B[\phi_<(x)] \\
    &= 1+U_{k_1 k_2qq} He_2(\phi_{k_1}(x),\phi_{k_2}(x))\\ 
    &=1+U_{k_1 k_2 qq} \left(\phi_{k_1}(x)\phi_{k_2}(x)-\delta_{k_1,k_2}\right)~,
\end{aligned}
\label{eq:Bflow}
\end{equation}
cf. \eqref{eq:trick}, shows that $\int\!\mathrm{d}\mu_x (1+2B(x))=1$ since $B$ is the second cumulant.

Interestingly, this result shows that the spatial weighting of the $L^2$ norm undergoes an RG flow. Specifically, after marginalizing over a single mode, we obtain 
\begin{equation}
	\frac{W_{\delta c}(x)}{\sigma^2_{\delta c}} = \frac{W_0(x)}{\sigma_0^2}-\frac{\delta c (1+2B_0(x))}{\sigma_0^4}
	\label{eq:WRG}
\end{equation}
where we introduce the notation $W_c(x)$ to represent this weighting, and set this to $1$ at the microscopic (UV) scale, $W_0(x)=1$\footnote{There is no redundancy between $W_c(x)$ and $\sigma_c^2$, as we require $\int\!\mathrm{d}\mu_x W_c(x)=1$.}. This is analogous to the renormalization of the mass in standard RG, where one defines a renormalized mass as a function of the energy scale; here, the renormalized coefficient of the quadratic term is denoted $W$, and depends on the data $x$.
Similarly, we denote by $B_0(x)$ the cumulant induced by integrating out the first mode. Harmoniously, we observe that integrating \eqref{eq:WRG} with respect to the data measure $\mathrm{d}\mu_x$ yields the previous RG flow for the ridge parameter, i.e., $\sigma_{\delta c}^2=\sigma_0^2+\delta c$. Plugging this in then yields the following equation for $W_{\delta c}(x)$:
\begin{equation}
	\frac{W_{\delta c}(x)}{\sigma^2_{0}+\delta c} = \frac{W_0(x)}{\sigma_0^2}-\frac{\delta c [1+2B_0(x)]}{\sigma_0^4}~,
\end{equation}
and hence
\begin{equation}
	W_{\delta c}(x) = W_0(x) - \frac{2 \, \delta c}{\sigma_0^2} B_0(x) + O\left(\delta c^2/\sigma^4_0\right)~,
\end{equation}
where we used the initial condition $W_0(x)=1$; notably, $W_{\delta c}(x)$ remains normalized. 

Intuitively, the weighting factor is quite natural: it reflects the fact that if, for example, the square of $\phi^2_q(x)$ has some positive correlations with $\phi^2_k(x)$ (where $q$ is the higher mode removed at a given RG step, and $k$ is some lesser mode), the marginalization results in varying amounts of effective noise, since more will be generated at spatial locations where large $\phi^2_k(x)$ is large.

\subsection{\label{sec:frg} Functional RG flow}

Extending the above single-mode integration into a potential flow equation requires handling two complications. The first is the need to express the distribution of features $P[\varphi_<]$ which was so far left abstract. To this end, we should  marginalize over $\phi_q$ in $P[\varphi]$. Conveniently, within the Edgeworth expansion, this marginalization amounts to omitting the $\varphi_q$ mode. This can be seen from \eqref{eq:Edge} and the fact that $He_{4}$ with any number of fixed lesser arguments integrates to zero under the greater Gaussian distribution. For instance, given one lesser argument and three greater arguments, we have  
\begin{equation}
\begin{aligned}
    \int\!&\mathrm{d}\varphi_q\,\mathcal{N}[0,1;\varphi_q] He_4[\varphi_{k},\varphi_{q},\varphi_{q},\varphi_q] \\
    &= \int\!\mathrm{d} \varphi_q\,\mathcal{N}[0,I;\varphi_k]^{-1}\partial_{\varphi_{k}}\mathcal{N}[0,I;\varphi_k]\,\partial_{\varphi_{q}}^3\mathcal{N}[0,I;\varphi_q] \\
    &=\varphi_k\!\int\!\mathrm{d}\varphi_q\left(\varphi_q^3-3\varphi_q\right)\mathcal{N}[0,I;\varphi_q] = 0~.
\end{aligned}
\end{equation}

The second issue is the appearance of a new structure -- the reweighting of the $L^2$ norm -- in the one-step renormalized action, which necessitates its inclusion in further RG steps. To this end, we repeat the previous derivation in its presence but take $y_>(x)=0$ for simplicity.  First, we write $W(x)=1+\delta W(x)$, and take the following ansatz for the infinitesimal change
\begin{equation}
	\delta W(x) = \sum_{k_1,k_2} W_{k_1 k_2} He_2[\phi_{k_1}(x),\phi_{k_2}(x)]~, 
\end{equation}
where based on (\ref{eq:WRG}) we have that at the second RG step we are currently carrying $W_{k_1 k_2} = -\frac{\delta c}{\sigma_0^2} U_{k_1 k_2 q q}$. We shall see below (\ref{Eq:WRG2nd}) that our RG flow respects this ansatz. Notably, for the current RG step, we denote the greater mode by $q'$. As $\delta W(x)$ contains pairs of features ($\varphi$'s) we split it into its lesser and greater components 
\begin{equation}
\begin{aligned}
    \delta W_<(x) &= \sum_{k_1,k_2 \neq q'} W_{k_1 k_2} He_2[\varphi_{k_1},\varphi_{k_2}]~, \\ 
    \delta W_>(x) &= 2\sum_{k_1 \neq q'} W_{k_1 q'} He_2[\varphi_{k_1},\varphi_{q'}] +W_{q',q'} He_2[\varphi_{q'},\varphi_{q'}]~.
\end{aligned}\label{eq:splitysplit}
\end{equation}
Following this, we may decompose the interaction term into lesser/greater components as
\begin{equation}
\begin{aligned}
    S_\mathrm{int}   =&\int\!\mathcal{D}\varphi\,P[\varphi_<,\varphi_>]\,\nonumber  \exp\!\left\{-\frac{1+\delta W_<+\delta W_>}{2 \sigma^2}\left(\GPl^\top\GPl+\GPh^\top\GPh+2\GPl^\top\GPh\right)\right\}~ \nonumber\\
    =&\int\!\mathcal{D}\varphi_< \,P[\varphi_<] \exp\!\left\{-\frac{1+\delta W_<}{2 \sigma^2}\GPl^\top\GPl\right\} \int\!\mathcal{D}\varphi_> P[\varphi_>|\varphi_<] \nonumber\\
    & \times\exp\!\left\{-\frac{1+\delta W_<+\delta W_>}{2 \sigma^2}\left(\GPh^\top\GPh+2\GPl^\top\GPh\right) -\frac{1}{2 \sigma^2}\delta W_>\GPl^\top\GPl\right\}~,
    \end{aligned}
\label{eq:newsplit2}
\end{equation}
cf. \eqref{eq:strawberry}.

Consistent with our earlier derivation, we aim to keep the leading order contribution in $U$, the non-Gaussian correction. We may thus expand the above to linear order in $\delta W_{\lessgtr}$. As before (cf. \eqref{eq:potato}), we also drop the $\GPh^\top \GPh$ term, thus leading to
\begin{equation}
\begin{aligned}
S_\mathrm{int} =&\int\!\mathcal{D}\varphi_< \,P[\varphi_<] \exp\!\left\{-\frac{1+\delta W_<}{2 \sigma^2}\GPl^\top\GPl\right\} \\
&\times \int\!\mathcal{D}\varphi_> P[\varphi_>|\varphi_<]\exp\!\left\{-\frac{1+\delta W_<}{ \sigma^2}\GPl^\top\GPh\right\}\left(1- \frac{\delta W_>}{\sigma^2}\GPl^\top\GPh - \frac{\delta W_>}{2\sigma^2}\GPl^\top\GPl  \right)~.
\end{aligned}
\label{eq:newsplit3}
\end{equation}
We next observe that if we take $\delta W_>=0$, the previous derivation could be repeated as-is under the replacement $\frac{1}{\sigma^2}\rightarrow \frac{1+\delta W_<}{\sigma^2}$. This is because $\delta W_<$ is a constant with respect to both the greater feature mode integration and the greater GP modes, and thus acts as a parameter under all algebraic manipulations carried out in the previous section. 

Our emphasis thus turns to the two $\delta W_>$ dependent terms. As these are linear order in $U$ (the non-Gaussian correction), under our leading order in $U$ treatment, we may evaluate their contribution while taking $\delta W_<=0$ (since $\delta W$ is $O(U)$) and $P[\varphi_>|\varphi_<]={\cal N}[0,I;\varphi_>]$ (since corrections to Gaussianity are $O(U)$). Namely, we may simplify the $\delta W_>$ dependence of the $\int\!\mathcal{D}\varphi_>$ integral into 
\begin{equation}
-\int\!\mathcal{D}\varphi_> {\cal N}[0,1;\varphi_>]\left(1-\frac{1}{ \sigma^2}\GPl^\top\GPh+\frac{1}{2}\left[\frac{1}{ \sigma^2}\GPl^\top\GPh\right]^2\right) \left(\frac{\delta W_>}{2\sigma^2}\GPl^\top\GPh+\frac{\delta W_>}{2\sigma^2}\GPl^\top\GPl \right),
\label{eq:TheStuff}
\end{equation}
where in the first bracket we kept up to second order in the greater GP mode since, as seen before, higher terms would contribute as $\delta c$ to a power larger than $1$ following the greater GP mode integration. In the second bracket, we kept up to leading order in $U$. 

The above integral leads to five terms, whose contributions to the renormalized GP action are analyzed in appendix \ref{sec:w_greater_terms}. The crucial observation that each such contribution appears as $O(\delta W_>)O(\delta c/\sigma^2)$ and consequently, as $\delta W_>$ is itself $O(\delta c/\sigma^2)$--- of order $(\delta c/\sigma^2)^2$. Such contributions are negligible within our momentum-shell (i.e., leading order in $\delta c/\sigma^2$) treatment. Indeed $O(\delta c/\sigma^2)^2$ terms, even after being summed $c/\delta c$-times during the RG process (where $c$ is the total variance of the integrated out modes), would be $O(c/\delta c)O(\delta c^2)=O(\delta c)$ and vanish in the $\delta c\rightarrow 0$ limit. In contrast, the $O(\delta c)$ terms (``Wilsonian" terms) may contribute $O(c/\delta c)O(\delta c)=O(c)$ and hence remain finite as $\delta c \rightarrow 0$. 

The above considerations imply that we may ignore all greater components of the weighting term in the RG process, and assume $W(x)$ contains only lesser feature modes. With this major simplification, we can proceed as before, with $\GPl^T \GPh/\sigma_c^2$ replaced by $W_{<}\GPl^T \GPh/\sigma_c^2$. As mentioned above, since $W_<$ is constant with respect to all RG manipulations (i.e., it is independent of the greater feature modes and all GP modes), this amounts to just carrying it through as a factor accompanying $\sigma^{-2}$. We thus find
\begin{equation}
\begin{aligned}
    \frac{W_{2\delta c}(x)}{\sigma_{2\delta c}^2} =& \frac{W_{\delta c}(x)}{\sigma_{\delta c}^2} - \frac{\delta c W_{\delta c}(x)^2 (1+2 B_{\delta c}(x))}{\sigma^4_{\delta c}} \nonumber \\ 
    =&\,\frac{1-\frac{2\,\delta c}{\sigma^2_0}B_0(x)}{\sigma_{\delta c}^2} - \frac{\delta c \left(1-\frac{2\,\delta c}{\sigma^2_0}B_0(x)\right)^2 (1+2 B_{\delta c}(x))}{\sigma^4_{\delta c}} \nonumber \\ 
    =&\,\frac{1-\frac{2\,\delta c}{\sigma^2_0}B_0(x)}{\sigma_{\delta c}^2} - \frac{\delta c  (1+ 2B_{\delta c}(x))}{\sigma^4_{\delta c}} + O(\delta c^2)~.
\end{aligned}
\end{equation}
As before, integrating both sides with respect to $\mathrm{d}\mu_x$ recovers the previous RG flow of $\sigma^2_{2\delta c}$, namely $\sigma^2_{2\delta c}=\sigma^2_{\delta c}+\delta c$ (we assume without loss of generality that both integrated modes have the same $\delta c$). Following a similar algebra to the our first RG step described in the previous subsection, we find 
\begin{equation}
\begin{aligned}
	W_{2 \delta c}(x) =&\,W_{\delta c}(x) - \frac{2\, \delta c}{\sigma_{\delta c}^2} [B_{\delta c}(x)] + O(\delta c^2) \\
	=&\, W_0(x) - \frac{2\, \delta c}{\sigma_{\delta c}^2} [B_{\delta c}(x)+B_{0}(x)]~,
\end{aligned}
\label{Eq:WRG2nd}
\end{equation}
which again integrates to 1 under $\mathrm{d}\mu_x$ (i.e., normalization is preserved). This time, no new structure appeared in our RG step, hence we may iterate it $M$ times to obtain the following RG flow 
\begin{equation}
\begin{aligned}
\sigma^2_c &= \sigma^2_0 + c~,\\ 
W_c(x)     &= 1-\sum_{m=0}^{M-1} \delta c B_{m \delta c}(x)/\sigma_{c} \\
	   &\approx 1 - 2 \int_0^{c}\!\mathrm{d}c'\,B_{c'}(x)/\sigma_c^2~, \label{eq:RGFlowNonGaussian}
\end{aligned} 
\end{equation}

where $\delta c M = c$ and the integral is obtain in the limit $\delta c \rightarrow 0$. Note that the full, input-dependent ridge parameter is $W_c(x)/\sigma_c^2$, cf. \eqref{eq:WRG}; here, we have split this into the data independent component and the spatial reweighting.

\subsection{\label{sec:toy} A solvable toy example: single mode integration}

As a concrete application, here we consider perhaps the simplest setting with tractable non-Gaussian effects. However, unlike in the above examples where such effects were treated perturbatively, here we consider a natural setting relevant to fully connected networks, wherein the lesser feature modes largely determine the greater feature modes. Concretely, we consider $x \in \mathbb{R}$ with $\mathrm{d}\mu_x \sim {\cal N}[0,1]$, and a rank-2 kernel of the form 
\begin{equation}
	\begin{aligned}
		K(x,y) &= \lambda_1 x y + \lambda_{2} (x^2 - 1)(y^2-1)\\
		       &=\lambda_1 He_1(x)He_1(y)+\lambda_2 He_2(x)He_2(y)~,
	\end{aligned}
\end{equation}
where $He_n(x)$ is again the probabilist's Hermite polynomial,
\begin{equation}
	He_n(x)\coloneqq (-1)^ne^{\frac{x^2}{2}}\frac{\mathrm{d}^n}{\mathrm{d}x^n}e^{-\frac{x^2}{2}}~.
\end{equation}
In particular, $He_1(x)=x$ and $He_2(x)=x^2-1$. Given the Gaussian measure, $He_1(x)$, $He_2(x)$ can be shown to span the non-zero eigenspaces of the kernel with eigenvalues ${\lambda_1:=\lambda_<}$, ${\lambda_2:=\lambda_>}$ respectively. To connect with our formalism above, note that here we have only two modes, $k\in\{1,2\}$, which are furnished by the first two Hermite polynomials, i.e., $\phi_1(x)=He_1(x)$ and $\phi_2(x)=He_2(x)$, cf. \eqref{eq:eigensys}. 

Our task is to learn a target function, 
\begin{equation}
	y(x)=x^5-10x^3+15x=He_5(x)~.
\end{equation}
From a spectral bias perspective \cite{canatar}, the average predictor for this model is zero, since the target is orthogonal to all non-zero eigenspaces under the input measure\footnote{By this we mean that the inner product on the associated Hilbert space vanishes, $\int\!\mathrm{d}\mu_x\,K(x,y) He_5(x)=0$.}. This is not necessarily the case however, since the features ($\varphi$, defined in \eqref{eq:trick}) will be non-Gaussian unless $\phi(x)$ is an affine function of $x\sim\mathcal{N}[0,1]$. Statistically therefore, the more common events in which $x$ lies closer to zero may bias the predictor on small datasets to act as if it were seeing the function $-10x^3+15x$, which does have non-vanishing overlap with $He_1(x)$. We show below that our RG approach can accurately predict such effects, and that they can be viewed as stemming from an MSE loss weighted by $\left(1-\lambda_2\sigma^{-2}He_2(x)^2\right)$. Indeed, while $y=He_5(x)$ and $He_1(x)$ are not coupled by standard MSE loss, the extra weighting term yields a non-zero overlap via integrals of the type $\int\!\mathrm{d}\mu_x He_1(x)\left(\lambda_2\sigma^{-2}He_2(x)^2\right) He_5(x)$. 

We shall first derive the above weighted form of the MSE loss, and then discuss its effect on predictions. Since the kernel mode associated with $He_5(x)$ is frozen/non-fluctuating, we do not treat it as a GP mode, and instead integrate out only the mode associated with $\lambda_2$\footnote{That is, we view $\bm y_<(x)$ as $\bm y_1He_1(x)+\bm y_5He_5(x)=He_5(x)$, with $y_1=0$ and $y_5=1$, and similarly $f_<(x)=f_1x$ with $f_5=0$. Alternatively, one may treat $y$ as part of the greater target modes, where its effect would then enter as a subleading shift to the lesser target.}. We denote this by $\varphi_>\eqqcolon\varphi_2$, and the associated replica-space GP mode by $\bm f_2$, with $\bm y_2 = 0$ (i.e., $q=2$ in our notation above). Furthermore, we take ${\bm f}_1$ to be the replica space vector associated with the linear (lesser) mode and (i.e., $He_1(x)$) and ${\bm y}_1$ to be the corresponding target. Converting to feature variables $\varphi_<\eqqcolon\varphi_1$ and $\varphi_>\eqqcolon\varphi_2$ via \eqref{eqn:varphilessgreat}, we have
\begin{equation}
\begin{aligned}
    P[\varphi_1] &= \int\mathrm{d}\mu_x\delta\left(\varphi_1-x\right)= \mathcal{N}[0,\mathbb{1};\varphi_1]~,\\
    P[\varphi_1,\varphi_2] &= P[\varphi_1]P[\varphi_2|\varphi_1] \nonumber\\
    &=\mathcal{N}[0,\mathbb{1};\varphi_1]\int\!\mathrm{d}\mu_x\delta\left(\varphi_2-He_2(x)\right) \\
    &=\mathcal{N}[0,\mathbb{1};\varphi_1]\delta\left(\varphi_2-He_2(\varphi_1)\right) \\
    &=\mathcal{N}[0,\mathbb{1};\varphi_1]\delta\left[\varphi_2-\left(\varphi_1^2-1\right)\right]~.
\end{aligned}
\label{eq:kartoffel}
\end{equation}
Notably, unlike the previously discussed perturbative case where $\phi_>$ was weakly affected by $\phi_<$, here $\phi_<$ fully determines $\phi_>$. (This scenario is relevant for real-world networks, since the same deterministic behavior is expected for any fully connected network with $\mathrm{d}\mu_x$ as the uniform measure on the hypersphere, where the lesser linear modes would determine all greater modes.) 

Now, consider the interaction term in the replicated partition function from \eqref{eq:newsplit}. As a result of the delta function, the integral over the greater feature mode can be easily resolved to yield
\begin{widetext}
\begin{equation}
\begin{aligned}
S_\mathrm{int} &=-\eta\int\!\mathrm{d}\varphi_1 \,P[\varphi_1] e^{-\frac{1}{2 \sigma^2}\Phi_1^\top\Phi_1}\int\!\mathrm{d}\varphi_2 P[\varphi_2|\varphi_1]\exp\!\left\{-\frac{1}{2 \sigma^2}\left(\Phi_2^\top\Phi_2+2\Phi_1^\top\Phi_2\right)\right\}\\
&=-\eta\int\!\mathrm{d}\varphi_1 \,P[\varphi_1] e^{-\frac{1}{2 \sigma^2}\Phi_1^\top\Phi_1}\int\!\mathrm{d}\varphi_2\delta\left(\varphi_2-(\varphi_1^2-1)\right)e^{-\frac{1}{2\sigma^2}\left(\varphi_2^2|\bm f_2|^2+2\varphi_2\varphi_1\bm f_2^\top(\bm f_1-\bm y_1)\right)}\\
&=-\eta\int\!\mathrm{d}\varphi_1 \,P[\varphi_1]\exp\left\{-\frac{1}{2 \sigma^2}\left(\Phi_1^\top\Phi_1+He_2(\varphi_1)^2|\bm f_2|^2 +2He_2(\varphi_1)He_1(\varphi_1)\bm f_2^\top(\bm f_1-\bm y_1)\right)\right\}.
\end{aligned}
\end{equation}
\end{widetext}
Next, we must integrate the greater GP mode $f_2$ in the replicated partition function \eqref{eq:path2}, $\left<Z^M\right>_\eta=e^{-\eta}\int\!\mathcal{D}\bm f\,e^{-S_0[\bm f]-S_\mathrm{int}}$. As in \eqref{eq:huge}, we capitalize on the fact that $f_2$ scales as $\sqrt{\lambda_2}$, and keep all contributions below order $f_2^4=O(\lambda_2^2)$. After expanding both exponentials, performing the integral, and re-exponentiating, we again find that the effect is to replace $f_2$ by the variance under $S_0[\bm f_2]$\footnote{Here we have also dropped the term $M\frac{\lambda_q}{2\sigma^2}He_2(\varphi_1)^2$ that arises from $|\bm f_q|^2=\mathrm{tr}(\bm f_q^\top \bm f_q)\rightarrow M\lambda_q$, since this vanishes in the replica limit. Such a factor of $M$ did not arise in \eqref{eq:huge}, since there $\left(\GPl^\top\bm f_q\right)^2\sim \bm f_q\bm f_q^\top$ does not involve a trace over replica indices.},
\begin{equation}
\begin{aligned}
\left<Z^M\right>_\eta&=\int\!\mathcal{D}f_1\exp\bigg\{-S_0[\bm f_1]+\eta\int\!\mathrm{d}\varphi_1P[\varphi_1]e^{-\frac{1}{2\sigma^2}\Phi_1^\top\Phi_1\left(1-\frac{\lambda_2}{\sigma^2}He_2(\varphi_1)^2\right)}\bigg\} \\
&=\int\!\mathcal{D}f_1\exp\bigg\{-S_0[\bm f_1]+\eta\int\!\mathrm{d}\mu_x\,e^{-\frac{1}{2\sigma^2}\Phi_1^\top\Phi_1\left(1-\frac{\lambda_2}{\sigma^2}He_2(x)^2\right)}\bigg\}~,
\end{aligned}
\label{eq:toyRG}
\end{equation}

where on the second line we used \eqref{eq:kartoffel} to replace $\varphi_1$, $\int\!\mathrm{d}\varphi_1P[\varphi_1]$ with $x$, $\int\!\mathrm{d}\mu_x$, respectively. We thus observe the advertised reweighting of the MSE term, namely
\begin{equation}
	\frac{1}{2\sigma^2}
	\;\;\overset{\textrm{RG}}{\longrightarrow}\;\;
	\frac{1}{2\sigma^2}\left(1-\frac{\lambda_2}{\sigma^2}He_2(x)^2\right)~.
	\label{eq:toyMSE}
\end{equation}
Next, we perform an EK-type approximation on \eqref{eq:toyRG}. Namely, we assume $\sigma^2$ is large enough compared to $\left(\bm f_1(x)-\bm y_1(x)\right)$ to perform a Taylor expansion of the exponent appearing in the action. We note in passing that the next-to-leading order in this expansion \cite{Cohen_2021} happens to vanish for $He_{n\geq 5}(x)$ (but not, e.g., for $He_3(x)$). Within the leading order expansion, the action in \eqref{eq:toyRG} becomes
\begin{equation}
S_0[\bm f_1] + S_\mathrm{int} = \frac{1}{2\lambda_1}\bm f_1^2 -\eta\int\mathrm{d}\mu_x\left(1-\frac{1}{2\sigma^2}(\bm f_1 x - He_5(x))^2 \left[1-\frac{\lambda_2}{\sigma^2} He_2(x)^2\right]\right)~.
\end{equation}
Without the weighting factor, the extremum of $f_1$ would be zero here as $He_1(x)=x$ and $He_5(x)$ do not overlap under $\int\mathrm{d}\mu_x$ (since only the cross-term, which is linear in $\bm f_1$, affects the average). However, with the weighting factor, and using $\int\mathrm{d}\mu_x\,x He_5(x) He_2(x)^2 = 120$, the saddlepoint is now
\begin{equation}
	\bar f_1
	\approx-\frac{\lambda_1}{\lambda_1+\sigma^2/\eta}\frac{120\lambda_2}{\sigma^2}~,
\end{equation}
where the approximation denotes leading order in $\lambda_2\sim\delta c$.

\begin{figure}[ht] 
\includegraphics[width=\textwidth, center, trim={0 0 0 1.1cm}, clip]{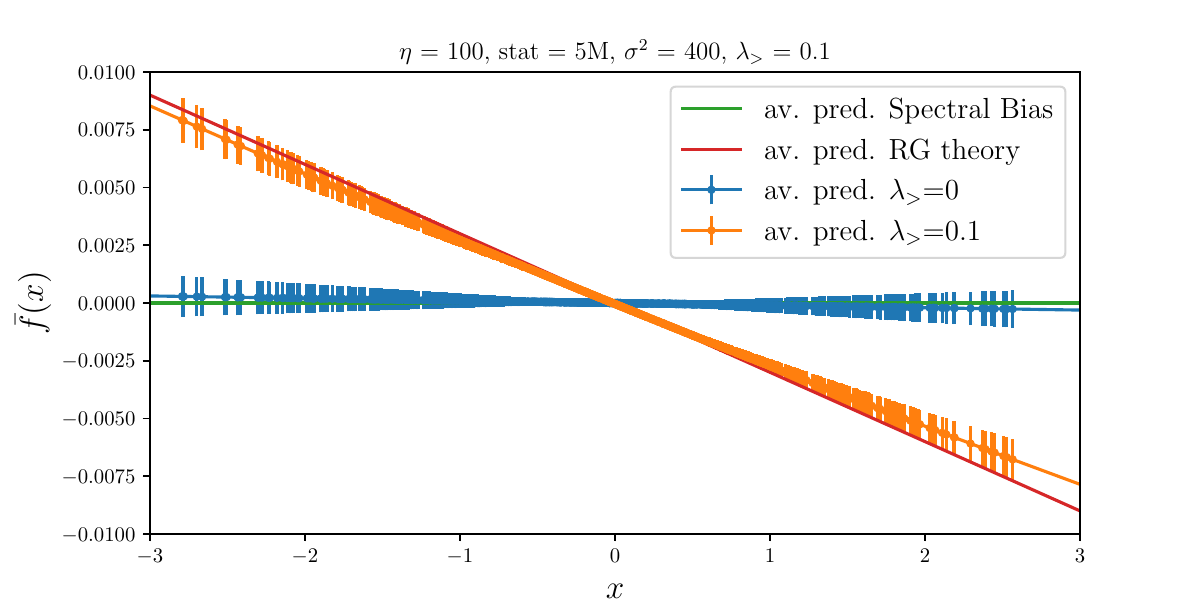}
\caption{\label{fig:empirical_results} {\bf Non-Gaussian features and spatial re-weighting effects}. Theory versus experiment for the model of sec. \ref{sec:toy} with $n=100$ datapoints $\sigma^2=400$ and $\lambda_>:=\lambda_2=0.1$ (unless stated otherwise). Learning a 5th Hermite polynomial, using a kernel capable of expressing only 1st and 2nd Hermite polynomials should give a zero average predictor (green line) based on the standard theory \cite{canatar,Cohen_2021}. However, due to spatial re-weighting, a coupling between 1st and 5th Hermite polynomial arises leading to a non-zero result.  For both $\lambda_>=0$ and $\lambda_>=0.1$, $m = 5$ million trials are performed. The average and standard error (i.e. standard deviation/$\sqrt{m}$) are reported.}
\end{figure}

To test this empirically, we sampled various draws of datasets with $n=100$ from $x \sim {\cal N}[0,1]$, performed GP regression on these with the above kernel using a fixed random test set with $n_\mathrm{test}=1000$, and averaged the results over the various draws (see fig.~\ref{fig:empirical_results}). Doing so for $\lambda_2=0$ shows an almost zero outcome, as expected. However, taking finite $\lambda_2$ reveals a trend that spectral bias \cite{Cohen_2021,canatar} misses. Notably, the fact that our toy example involves only one mode which is integrated out, together with the need to take its variance to be small  ($\lambda_2/\sigma^2 \ll 1$) as well as make it unlearnable ($\eta \lambda_2/\sigma^2 \ll 1$) results in a rather small effect. Still, it serves to demonstrate the conceptual phenomena whose importance depends on the setting or the desired accuracy. 

\section{\label{sec:discussion} Discussion}

In this work, we applied the powerful formalism of Wilsonian RG to Gaussian Process regression. In particular, treating the output of a trained DNN as a statistical ensemble, we split the associated fluctuations into high- and low-energy modes (based on their stiffness in a random network). Provided the variance of the higher modes is much smaller than the ridge parameter, these may be integrated out in a tractable manner leading to a renormalized GP regression problem. In simple settings where the features associated with those modes are Gaussian, this results in a renormalized ridge parameter. We show that this simple setting can already achieve impressive results on real-world datasets. In particular, we show that our RG formalism can accurately predict the semi power-law scaling behavior of the MSE loss as a function of training dataset size on both MNIST and CIFAR10 regression tasks.‌ In more general settings, one obtains an effective GP regression problem where the ridge parameter becomes spatially dependent. We then applied this to a simple toy model, in which the effects determined via our RG approach are demonstrably missed in the standard eigen-learning framework \cite{canatar}.

It would be interesting to generalize our formalism to include higher order contributions in the Non-Gaussianity as well as fluctuation effects over datasets by keeping track of $y_>^2$ dependencies. Such an effort would also benefit from diagrammatic reformulation of our derivation. Such a formalism may enable us to define a notion of relevancy and irrelevancy applicable to generic non-Gaussian features and, potentially, a notion of universality classes for DNNs. Similarly, our formalism could be extended to finite neural networks where additional interaction terms would appear corresponding to finite-width corrections to the Gaussian prior, and investigate universality in those corrections.  

\begin{acknowledgments}
We wish to thank Marc Klinger and Alexander Stapleton for useful discussions regarding the Bayesian Renormalization Group and related topics.
JNH was supported by the National Science Foundation under Grant No. NSF
PHY-1748958 and by the Gordon and Betty Moore Foundation through Grant No. GBMF7392. AM acknowledges support from Perimeter Institute, which  is supported in part by the Government of Canada through the Department of Innovation, Science and Economic Development and by the Province of Ontario through the Ministry of Colleges and Universities. ZR acknowledges support from ISF Grant 2250/19. JNH, RJ, and AM wish to thank the Mainz Institute for Theoretical Physics for hosting the 2022 workshop ``A Deep-Learning Era of Particle Theory'', where this collaboration was initiated, as well as the Aspen Center for Physics (ACP, supported by National Science Foundation grant PHY-2210452) for hospitality during the 2023 winter conference ``Theoretical Physics for Machine Learning''. JNH, and ZR acknowledge further support from ACP through the 2023 summer program ``Theoretical Physics for Deep Learning''. This research was also supported in part by grants NSF PHY-1748958 and PHY-2309135 to the Kavli Institute for Theoretical Physics (KITP), through the participation of AM, JNH, and ZR in the Fall 2023 program ``Deep Learning from the Perspective of Physics and Neuroscience.'' 
\end{acknowledgments}

\appendix
\section{\label{sec:spectral} Spectral decomposition of the Gaussian action}

In this appendix, we perform the spectral decomposition given in \eqref{eq:specdecomp} on the quadratic action \eqref{eq:quadS}, repeated here for convenience:
\begin{equation}
	S = \sum_{m=1}^M \frac{1}{2}\int\!\mathrm{d}\mu_x \mathrm{d}\mu_{x'}\,f_m(x) K^{-1}(x,x') f_m(x') 
           + \frac{\eta}{\sigma^2} \int\!\mathrm{d}\mu_x \sum_m \frac{\left(f_m(x)-y(x)\right)^2}{2}~.
\label{eq:Squadrep}
\end{equation}
Beginning with the first term, we have
\begin{equation} 
\begin{aligned}
    \int\!&\mathrm{d}\mu_x \mathrm{d}\mu_{x'}\,f_m(x) K^{-1}(x,x') f_m(x')\\
          &= \int\!\mathrm{d}\mu_x \mathrm{d}\mu_{x'} \left( \sum_k f_{mk} \phi_k(x) \right) K^{-1}(x,x') \left( \sum_{k'} f_{mk'} \phi_{k'}(x') \right)\\
          &= \sum_{k,k'} f_{mk} f_{mk'}\!\int\!\mathrm{d}\mu_x \phi_k(x) \left( \int\!\mathrm{d}\mu_{x'}  K^{-1}(x,x') \phi_{k'}(x') \right)\\
          &= \sum_{k,k'} \lambda_{k'}^{-1}f_{mk} f_{mk'}\!\int\!\mathrm{d}\mu_x\,\phi_k(x)\phi_{k'}(x)\\
          &= \sum_{k,k'}\lambda_{k'}^{-1}f_{mk} f_{mk'}\delta_{kk'}
          = \sum_k \frac{1}{\lambda_{k}} |f_{mk}|^2~,
\end{aligned}
\label{eq:term1}
\end{equation}
where in going to the fourth line we have used \eqref{eq:eigensys}, and the last line follows from the orthonormality relation \eqref{eq:orthophi}. Meanwhile, in the second term we have
\begin{equation}
\begin{aligned}
     \int\!&\mathrm{d}\mu_x\left(f_m(x)-y(x)\right)^2 \\
	   &=\int\!\mathrm{d}\mu_x \left[ \sum_k \left( f_{mk} - y_k \right) \phi_k(x) \right]\left[ \sum_{k'} \left( f_{mk'} - y_{k'} \right) \phi_{k'}(x) \right] \\
        &= \sum_{k,k'} \left( f_{mk} - y_k \right) \left( f_{mk'} - y_{k'} \right)\!\int\!\mathrm{d}\mu_x\,\phi_k(x) \phi_{k'}(x) \\
        &= \sum_k \left( f_{mk} - y_k \right)^2~,
\end{aligned}
\end{equation}
where the last step again follows from \eqref{eq:orthophi}. Substituting these into \eqref{eq:Squadrep} then yields the decoupled feature-space action \eqref{eq:featureS}, as desired.

Note that these expressions will be unchanged when performing the split into greater and lesser GP modes in \eqref{eq:split1}, since the cross-terms will vanish by the orthonormality relation \eqref{eq:orthophi}. For example, the fourth line of \eqref{eq:term1} becomes
\begin{equation}
\begin{aligned}
    \int\!&\mathrm{d}\mu_x\left(\sum_{k\leq\kappa}f_{mk}\phi_k+\sum_{k>\kappa}f_{mk}\phi_k\right)\left(\sum_{k'\leq\kappa}\lambda_{k'}^{-1}f_{mk'}\phi_{k'}+\sum_{k'>\kappa}\lambda_{k'}^{-1}f_{mk'}\phi_k'\right)\\
    &=\left(\sum_{k,k'\leq\kappa}+\sum_{k,k'>\kappa}+\sum_{\substack{k\leq\kappa\\k'>\kappa}}+\sum_{\substack{k>\kappa\\k'\leq\kappa}}\right)\lambda_{k'}^{-1}f_{mk}f_{mk'}\int\!\mathrm{d}\mu_x\,\phi_k(x)\phi_{k'}(x)\\
      &=\sum_k\frac{1}{\lambda_k}|f_{mk}|^2
\end{aligned}
\end{equation}
where the last two summations vanish since $k\!\neq k'$.


\section{Applying the Woodbury identity}\label{sec:woodbury}
In this short appendix, we detail the application of the Woodbury matrix identity in \eqref{eq:WoodburyFF}. In particular, we apply the identity in the form
\begin{equation}
	\mathbb{1}_{M\times M}-U\left(\mathbb{1}_{N\times N}+VU\right)^{-1}V=\left(\mathbb{1}_{M\times M}+UV\right)^{-1}~,
	\label{eq:WoodID}
\end{equation}
where $U$ is $M\!\times\!N$ and $V$ is $N\!\times\!M$. Since the matrix ${C=FF^\top}$ defined in \eqref{eq:covmat} has dimensions $M\!\times\!M$, we identify $U=F/\sigma$ and $V=F^\top/\sigma$, whereupon \eqref{eq:WoodID} becomes
\begin{equation}
	\mathbb{1}-\frac{1}{\sigma^2}F\left(\mathbb{1}+F^\top F/\sigma^2\right)^{-1}F^\top=\left(\mathbb{1}+FF^\top/\sigma^2\right)^{-1}~.
\end{equation}
A trivial rearrangement yields
\begin{equation}
	\begin{aligned}	
		F&\left(\mathbb{1}+F^\top F/\sigma^2\right)^{-1}F^\top=\sigma^2-\sigma^2\left(\mathbb{1}+C/\sigma^2\right)^{-1}\\
		&=\sigma^2-\frac{\sigma^4}{\sigma^2+C}
		=\frac{\sigma^2 C}{\sigma^2+C}
		=\left(C^{-1}+\sigma^{-2}\right)^{-1}~,
	\end{aligned}
\end{equation}
which is \eqref{eq:WoodburyFF}.


\section{\label{App:NoY} Simplifying interactions in the replica limit}
The replicated partition function allows us to calculate observables by adding some source term (e.g., $\sum_{k} \alpha_k f_{k}$) and taking functional derivatives,
\begin{equation}
\partial_{\alpha} \langle \ln(Z_{\alpha})\rangle_{\eta} \mathrel{\mathop{=}\limits_{M \rightarrow 0}} \partial_{\alpha} \frac{\langle Z_{\alpha}^M \rangle_{\eta}-1}{M}~.
\end{equation}
Such a procedure has a predictable dependence on $y$. In particular, since each $\ln(Z_{\alpha})$ describes the Gaussian Process/distribution for a given training set, only $\partial_{\alpha} \ln(Z_{\alpha})$ scales as $y$, while higher derivatives/cumulants do not. 
This scaling survives dataset averaging. Consequently, we may imagine an infinitesimal scaling variable $\epsilon$, take $y\rightarrow \epsilon y$, and remember to divide by $\epsilon^{-1}$ the average predictor (but not any of the higher cumulants). Equivalently, we may ignore any $y^2$ or higher contribution, and use the following equivalent version of $C$:
\begin{equation}
	C_{mm'} \rightarrow \sum_{k > \kappa} \left( f_{km} f_{km'} - y_k (f_{km} + f_{km'})   \right)
	\label{eq:calt}
\end{equation}
cf. \eqref{eq:covmat}.

Our next aim is to show that perturbation theory in $C$, where $C$ appears to the $n^\mathrm{th}$ order, yields terms which scale as $\big(\delta c/\sigma_c^2\big)^n$. The power of $\sigma_c^2$ stems from the simple fact that $C$ always appears accompanied by $\sigma^2$ in the interaction, which becomes the effective parameter $\sigma_c^2$ under RG. Meanwhile, the power of $\delta c$ can be understood from the fact that, had we not included the $y_k$ component in $C_{mm'}$, the two factors of $f_{mk}$ from perturbation theory with respect to the free theory for the greater modes (cf. $S_0[f_>]$ in \eqref{eq:featureS}) would contract with the same GP mode, thereby yielding the desired power of $\delta c$. Turning to the $y_k$ term may then be treated as adding a $\sqrt{\delta c}$ factor, since the single factor of $f_{mk}$ appearing in \eqref{eq:calt} averages to zero and must get contracted. However, due to our argument about $y$ above, it may only appear to first power overall in the perturbative expansion (indeed higher power would yield a free replica summation which would nullify in the replica limit). Furthermore, being antisymmetric in the greater modes ($f_{mk}$), it may only appear when calculating observables with an odd power of greater modes. Subsequently, if we are interested only in lesser observables we may take $y_>=0$ in the derivation for Gaussian features.

Using the above reasoning and our previous infinitesimal (i.e., leading order $\delta c/\sigma^2$) approach, we may integrate out the greater GP modes using leading order perturbation theory in $C$. As mentioned in the main text, this essentially amounts to replacing $C$ by its average under the free theory wherever it appears in the interaction. As this yields a $\delta c \mathbb{1}_{M \times M}$ with no $y_>$ dependence, we may take $y_>=0$ from the start in the Gaussian case when calculating lesser observables. The effect of this procedure on the ``loss'' part of the interaction is discussed in the main text. This procedure also exposes that the prefactor $\exp\left\{-\frac{1}{2}\mathrm{tr}\ln\left(C/\sigma^{2}+\mathbb{1} \right)\right\}$ appearing in  \eqref{eq:newresult} becomes $\exp\left\{-\frac{1}{2}\mathrm{tr}\ln\left(\left(\delta c/\sigma^{2}+1\right)\mathbb{1}_{M\times M}\right)\right\}=\exp\left\{-\frac{M}{2}\mathrm{tr}\ln\left(\delta c/\sigma^{2}+1\right)\right\}$, which goes to 1 in the replica ($M\to 0$) limit. This justifies its removal in the main text. 

The same general reasoning applies to the non-Gaussian case, however extra care is needed for the terms proportional to $A$ as these contain a single power of $y$ which may affect the average predictor. 

\section{\label{sec:w_greater_terms} $\delta W_>(x)$ dependent terms}

To evaluate \eqref{eq:TheStuff}, it is convenient to split $\delta W_>$ into components involving a single greater feature mode, which we denote $\delta W_>^{(1)}$, and those involving two greater feature modes, which we denote $\delta W_>^{(2)}$, cf. \eqref{eq:splitysplit}. 

Now, consider first the term in \eqref{eq:TheStuff}, proportional to $\frac{\delta W^{(2)}_>}{\sigma^2}\GPl^\top\GPh$. 
As it is odd in the greater feature mode, we have 
\begin{equation}
\begin{aligned}
        -&\int\!\mathcal{D}\varphi_> {\cal N}[0,1;\varphi_>]\left(1-\frac{1}{ \sigma^2}\GPl^\top\GPh+\frac{1}{2}\left[\frac{1}{ \sigma^2}\GPl^\top\GPh\right]^2\right)\frac{\delta W^{(2)}_>}{2\sigma^2}\GPl^\top\GPh\\
         &= \int\!\mathcal{D}\varphi_> {\cal N}[0,1;\varphi_>]\frac{1}{2\sigma^4}\left(\GPl^\top\GPh\right)^2 \delta W^{(2)}_>\\
        &= \frac{1}{2\sigma^4}\left(\GPl^\top\fqt\right)^2\int\!\mathcal{D}\varphi_> {\cal N}[0,1;\varphi_>] \delta W^{(2)}_> \phi^2_{q'}\\
        &= \frac{1}{\sigma^4}\left(\GPl^\top\fqt\right)^2 W_{q'q'}
        \eqqcolon T_1~.
\end{aligned}	
\end{equation}
As we are keeping only the leading order in $W$, under the GP integration this term would simply be replaced by its average under the free greater mode action where it would yield the following additional contribution to the renormalized GP action  
\begin{equation}
C_1 \coloneqq \int\!\mathcal{D}\varphi_< \,P[\varphi_<] \exp\!\left\{-\frac{1+\delta W_<}{2 \sigma^2}\GPl^\top\GPl\right\} \frac{1}{\sigma^2}\GPl^\top \GPl \frac{W_{q'q'} \lambda_{q'}}{\sigma^2}~.
\label{eq:C1}
\end{equation}
Next, we consider the term proportional to $\frac{\delta W^{(1)}_>}{\sigma^2}\GPl^\top\GPh$. As it is even in the greater feature modes, we have 
\begin{equation}
	\begin{aligned}
		-&\int\!\mathcal{D}\varphi_> {\cal N}[0,1;\varphi_>]\left(1-\frac{1}{ \sigma^2}\GPl^\top\GPh+\frac{1}{2}\left[\frac{1}{ \sigma^2}\GPl^\top\GPh\right]^2\right)\frac{\delta W^{(1)}_>}{2\sigma^2}\GPl^\top\GPh\\
		 &=-\int\!\mathcal{D}\varphi_> {\cal N}[0,1;\varphi_>]\left(1+\frac{1}{2}\left[\frac{1}{ \sigma^2}\GPl^\top\GPh\right]^2\right)\frac{\delta W^{(1)}_>}{2\sigma^2}\GPl^\top\GPh\\
		&=-\frac{1}{2\sigma^2}\GPl^\top\fqt \int\!\mathcal{D}\varphi_> {\cal N}[0,1;\varphi_>] \delta W^{(1)}_> \varphi_{q'}-2\left[\frac{1}{2\sigma^2}\GPl^\top\fqt\right]^3 \int\!\mathcal{D}\varphi_> {\cal N}[0,1;\varphi_>] \delta W^{(1)}_> \varphi_{q'}^3 \\
		&=-\frac{1}{2\sigma^2}\GPl^\top\fqt \left(\sum_{k_1 \neq q'} W_{k_1,q'} \varphi_{k_1}\right)-2\left[\frac{1}{2\sigma^2}\GPl^\top\fqt\right]^3 \left(3\sum_{k_1 \neq q'} W_{k_1,q'} \varphi_{k_1}\right)\\
  & \eqqcolon T_2+T_3~.
	\end{aligned}
\end{equation}
Conveniently, the corresponding contribution ($C_2$, $C_3$, cf. \eqref{eq:C1}) to the renormalized GP action would be proportional to the average of these two terms under the free greater GP mode action, which vanishes due to their odd order in $\fqt$.

We next consider the term proportional to $\frac{\delta W^{(1)}_>}{2\sigma^2}\GPl^\top\GPl$ in \eqref{eq:TheStuff}. Using the fact that this is odd in the greater feature mode we have 
\begin{equation}
\begin{aligned}
		-\int\!&\mathcal{D}\varphi_> {\cal N}[0,1;\varphi_>]\left(1-\frac{1}{ \sigma^2}\GPl^\top\GPh+\frac{1}{2}\left[\frac{1}{ \sigma^2}\GPl^\top\GPh\right]^2\right) \frac{\delta W^{(1)}_>}{2\sigma^2}\GPl^\top\GPl\\
		&= \int\!\mathcal{D}\varphi_> {\cal N}[0,1;\varphi_>]\frac{1}{ \sigma^2}\GPl^\top\GPh\frac{\delta W^{(1)}_>}{2\sigma^2}\GPl^\top\GPl \\ 
		&= \frac{1}{ \sigma^2}\GPl^\top\fqt \frac{1}{2\sigma^2}\GPl^\top\GPl \int\!\mathcal{D}\varphi_> {\cal N}[0,1;\varphi_>] \delta W^{(1)}_> \varphi_{q'}\\
  &\eqqcolon T_4~.
\end{aligned}
\end{equation}
However, the associated contribution ($C_4$) to the renormalized GP action again vanishes due to the odd order in $\fqt$.

Finally, we consider the term proportional to $\frac{\delta W^{(2)}_>}{2\sigma^2}\GPl^\top\GPl$, which is even in the greater feature mode, hence
\begin{equation}
	\begin{aligned}
		-\int\!&\mathcal{D}\varphi_> {\cal N}[0,1;\varphi_>]\left(1-\frac{1}{ \sigma^2}\GPl^\top\GPh+\frac{1}{2}\left[\frac{1}{ \sigma^2}\GPl^\top\GPh\right]^2\right) \frac{\delta W^{(2)}_>}{2\sigma^2}\GPl^\top\GPl\\
		&=-\int\!\mathcal{D}\varphi_> {\cal N}[0,1;\varphi_>]\frac{1}{4\sigma^2}\left[\frac{1}{ \sigma^2}\GPl^\top\GPh\right]^2\delta W^{(2)}_>\GPl^\top\GPl \\ 
		&=-\frac{1}{4\sigma^2}\left[\frac{1}{ \sigma^2}\GPl^\top \fqt \right]^2\GPl^\top\GPl \int\!\mathcal{D}\varphi_> {\cal N}[0,1;\varphi_>] \delta W^{(2)}_> \varphi_{q'}^2 \\
		&= -\frac{1}{2\sigma^2}\left[\frac{1}{ \sigma^2}\GPl^\top \fqt \right]^2\GPl^\top\GPl W_{q' q'} \eqqcolon T_5~,
	\end{aligned}
\end{equation}
where the first term in the first line vanishes since $He_2(\phi_q^2)$ integrates to zero under the standard normal measure. This yields the following contribution to the renormalized GP action 
\begin{equation}
C_5\coloneqq-\int\!\mathcal{D}\varphi_< \,P[\varphi_<] \exp\!\left\{-\frac{1+\delta W_<}{2 \sigma^2}\GPl^\top\GPl\right\}
            \frac{1}{2}\left[\frac{1}{ \sigma^2}\GPl^\top \GPl \right]^2 \frac{W_{q' q'}\lambda_{q'}}{\sigma^2}~.
\end{equation}
However, both non-zero contributions ($C_1$, $C_5$) are proportional to $W \lambda_{q'}/\sigma^2$, and hence contribute at order $[\delta c/\sigma_c^2]^2$. Integrating out those modes up to a total variance of ${\sum^{k_{min}}_{k_{max}} \delta_c=c}$ following the renormalization flow \eqref{eq:RGFlowNonGaussian}, $W_{kk'}$ may be  $O(c)$ and non-negligible; however,  any contribution of the type $W_{kk'}\delta c$ would still need to vanish in the Wilsonian, $\delta c \rightarrow 0$, limit. 

\section{\label{app:real_data_vary_threshold} Effect of varying threshold $T$ in real-world data experiments}

\begin{figure}[ht]
    \centering
    \includegraphics[width=0.49\linewidth]{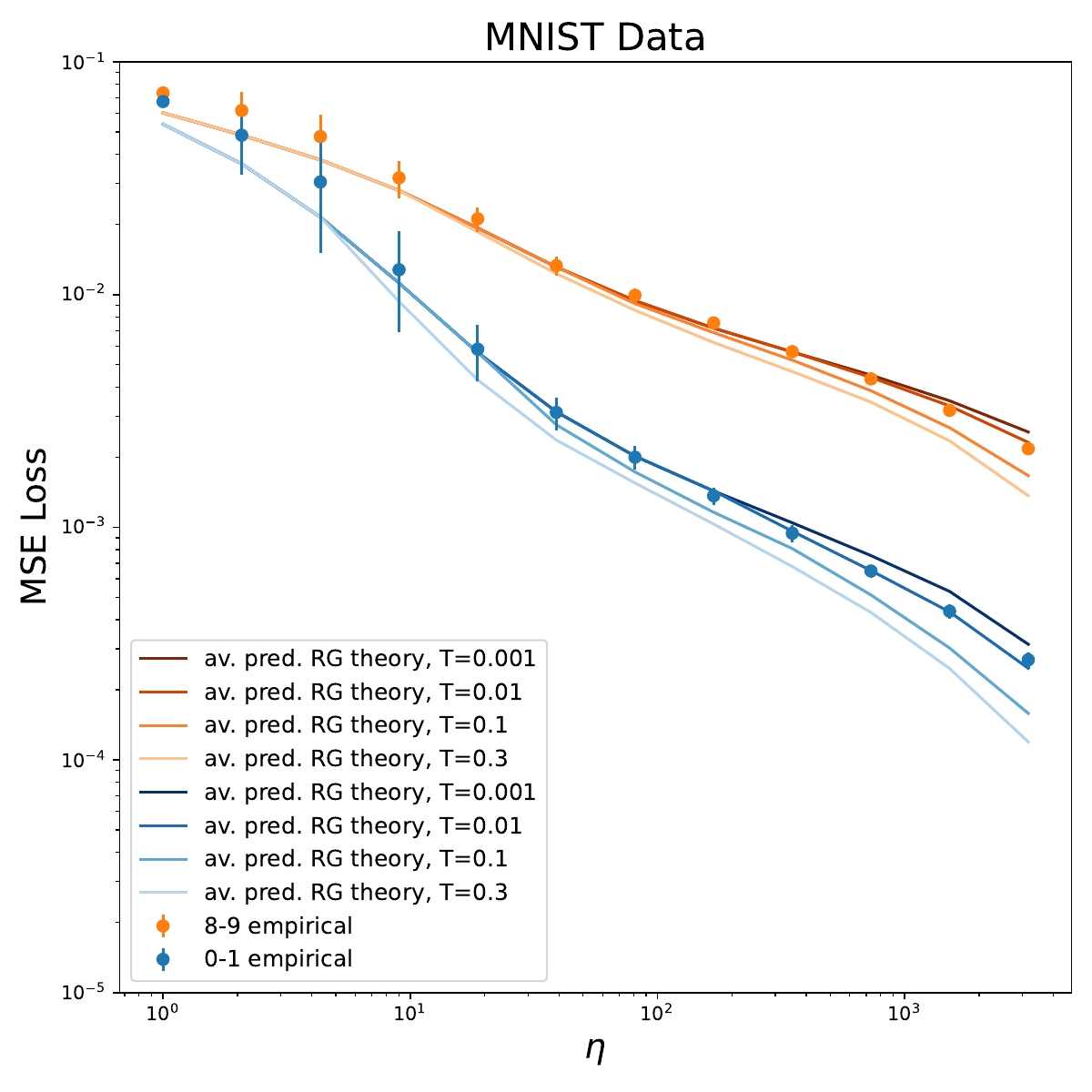}
    \includegraphics[width=0.49\linewidth]{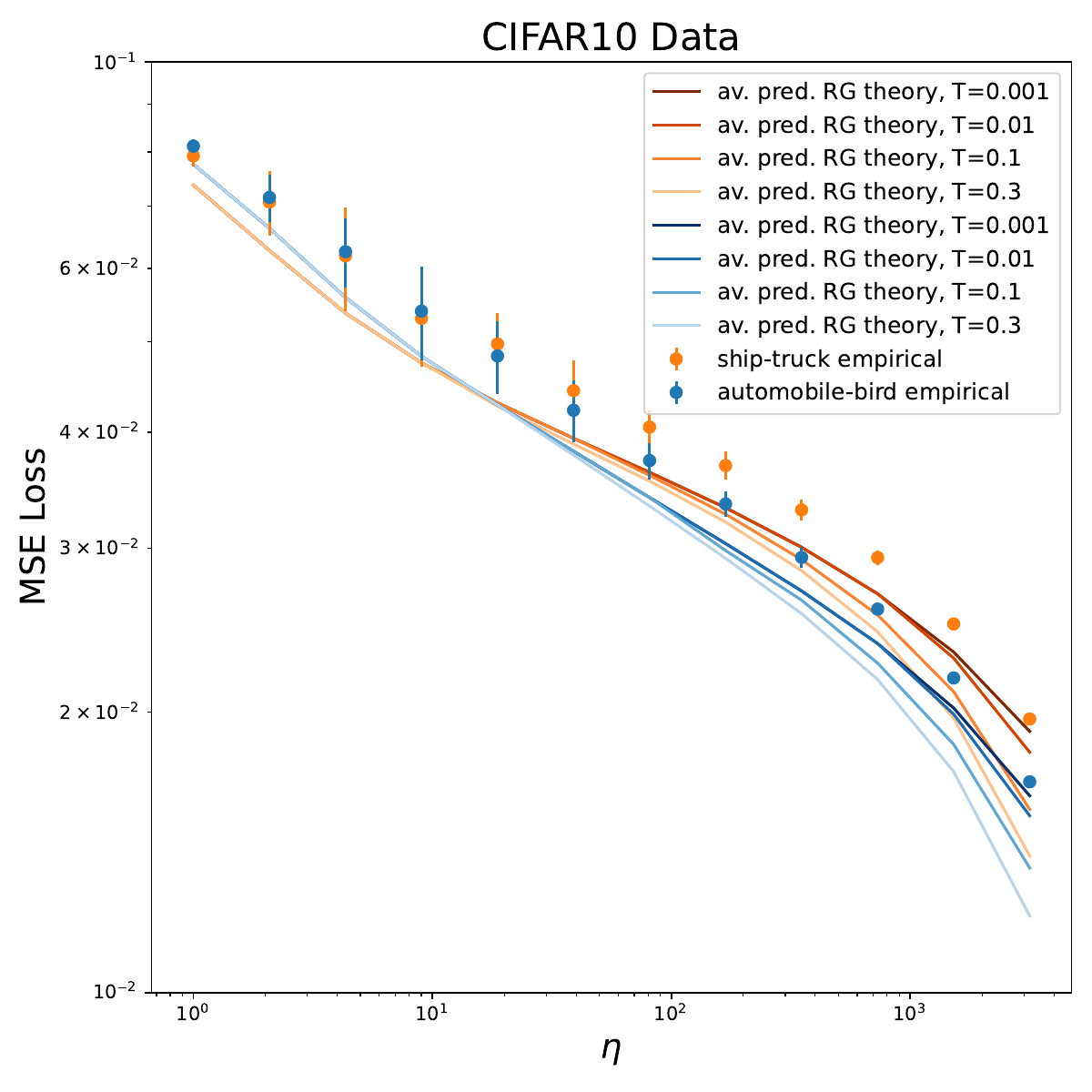}
    \caption{ \textbf{Effect of varying the learnability threshold, $T$.} For the two real-world dataset examples (MNIST and CIFAR10) considered in section ~\ref{sec:scalingLaws_empirical}, we show the effect of the choice of learnability threshold $T\in (0,1)$. The choice of $T$ inadvertently determines the learnability cutoff, $\kappa$, through finding $\kappa$ such that the learnability factor $L_\kappa \approx T$ (see equation~\ref{eq:learnabilityfactor}). While this choice slightly changes the RG theory prediction at high $\eta$, the effect is relatively minor.}
    \label{fig:empirical_results_scalinglaws_varythreshold}
\end{figure}

In this appendix we include additional plots in figure~\ref{fig:empirical_results_scalinglaws_varythreshold} that show the effect of the choice of threshold, $T$, on the numerical experiments presented in section~\ref{sec:scalingLaws_empirical}.

\section{\label{app:gaussian_vs_cauchy} Exploration of the distribution of feature modes as a function of $k$}

\begin{figure}
    \centering
    \includegraphics[width=0.49\linewidth]{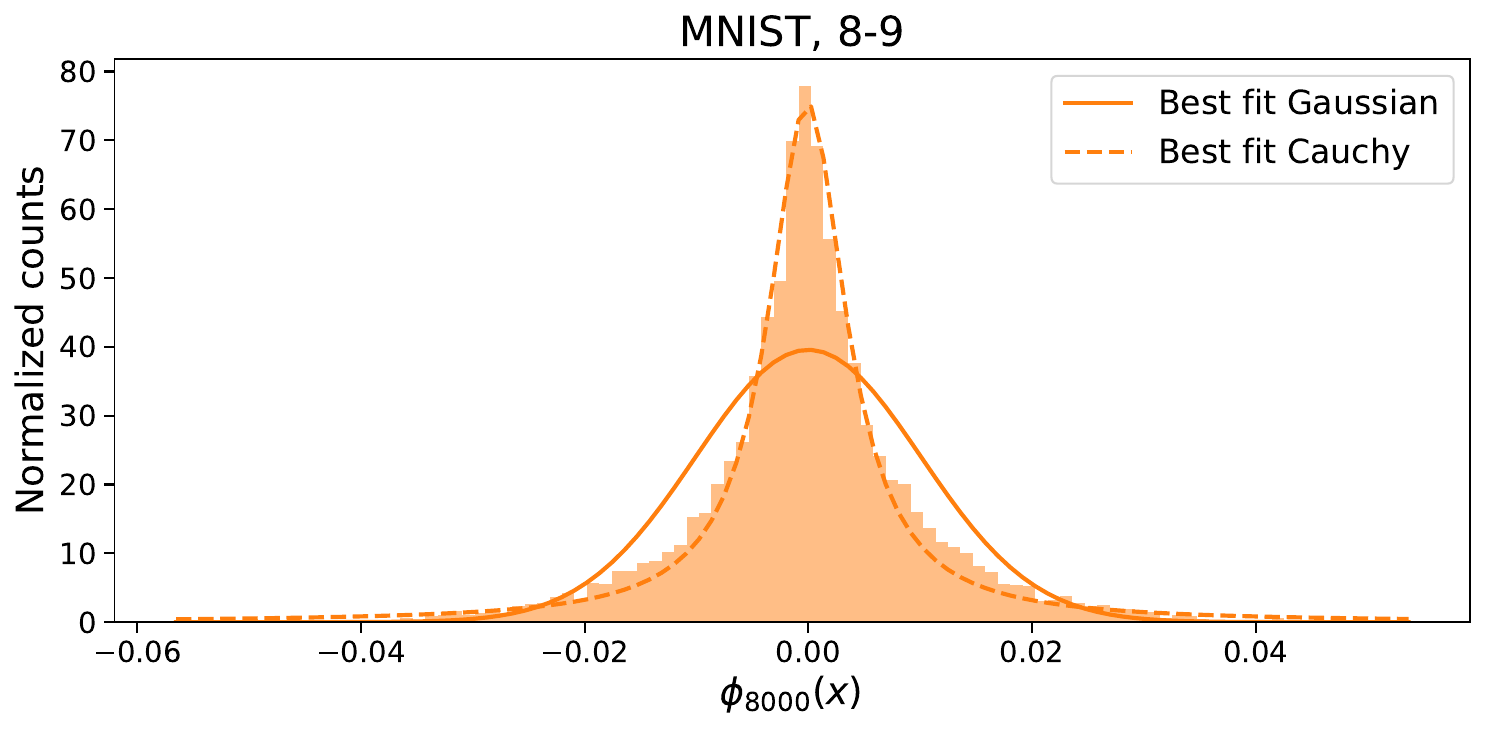}
    \includegraphics[width=0.49\linewidth]{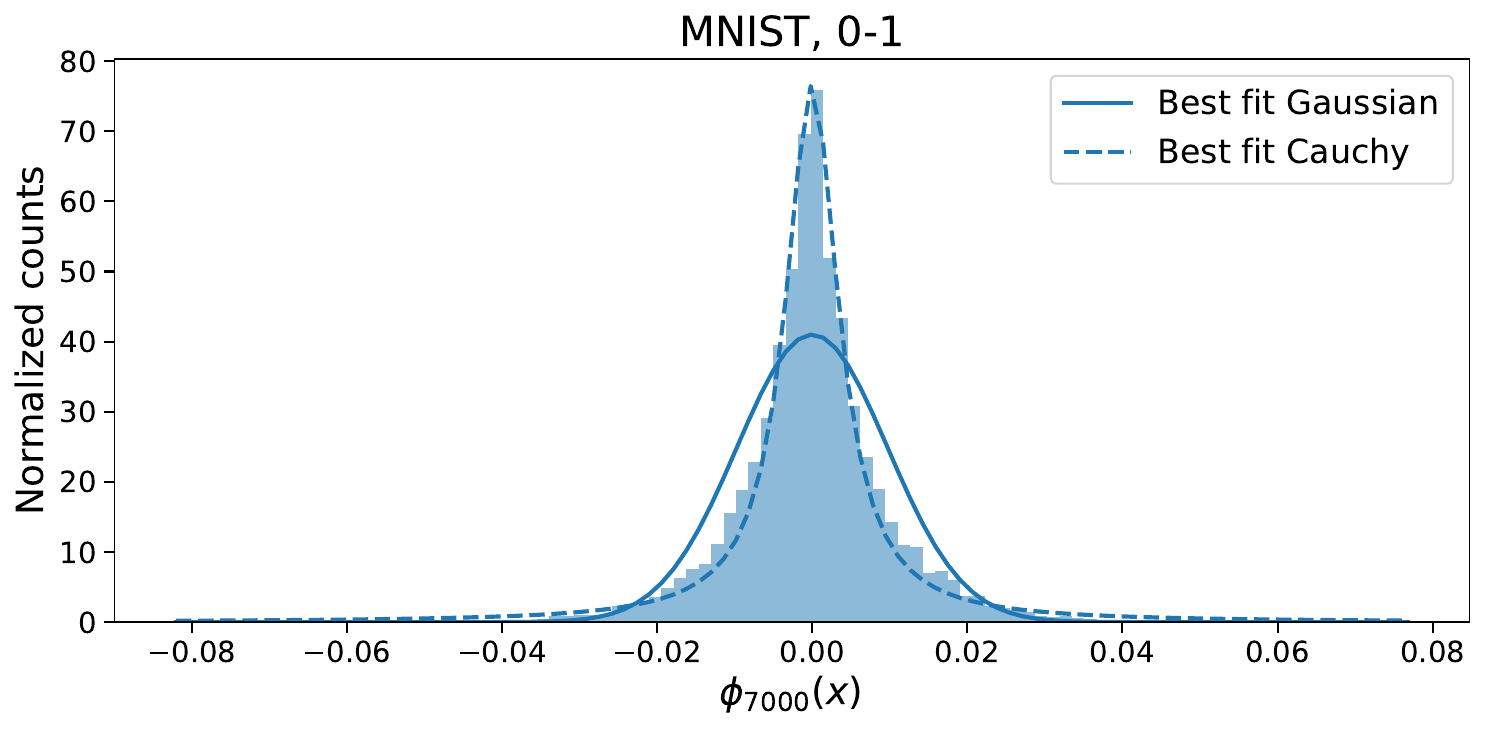}
    \includegraphics[width=0.49\linewidth]{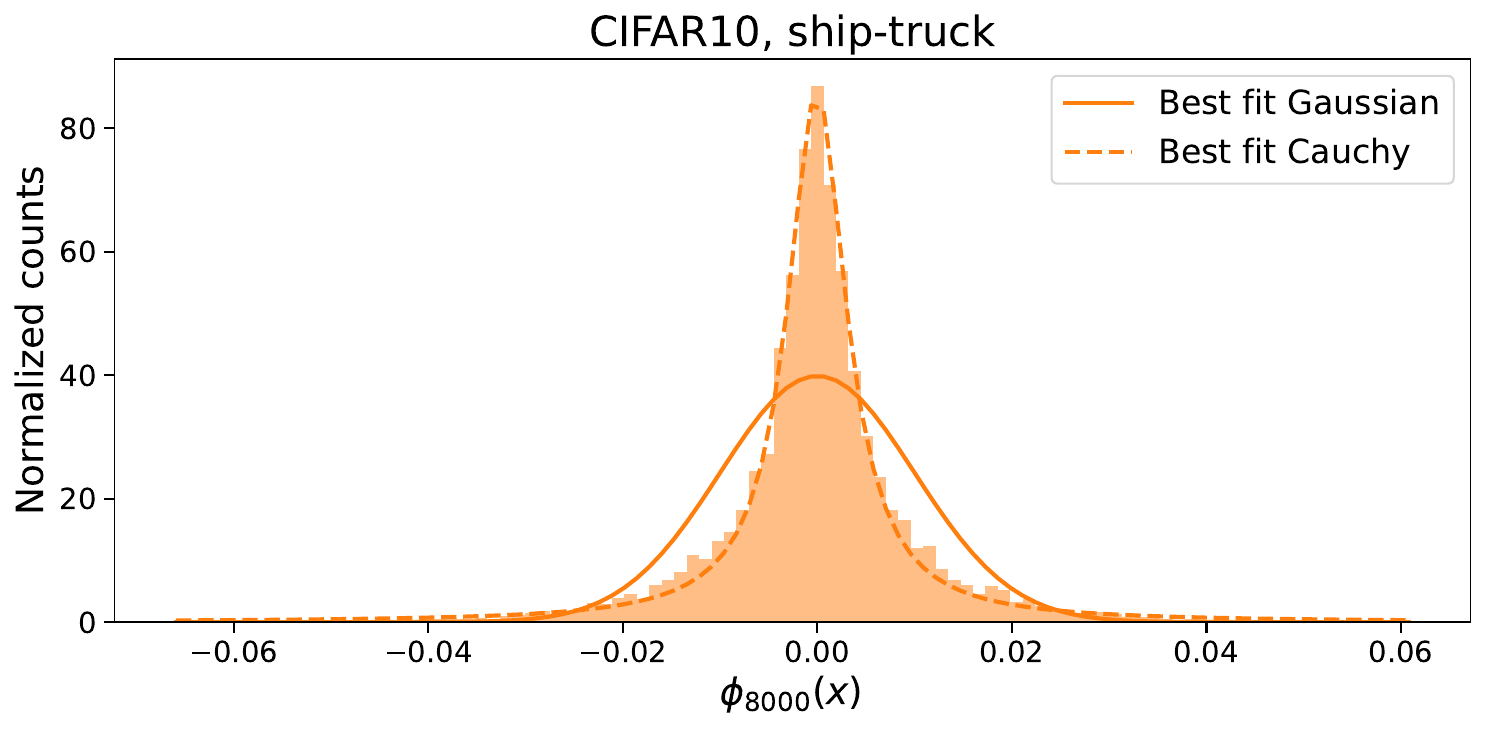}
    \includegraphics[width=0.49\linewidth]{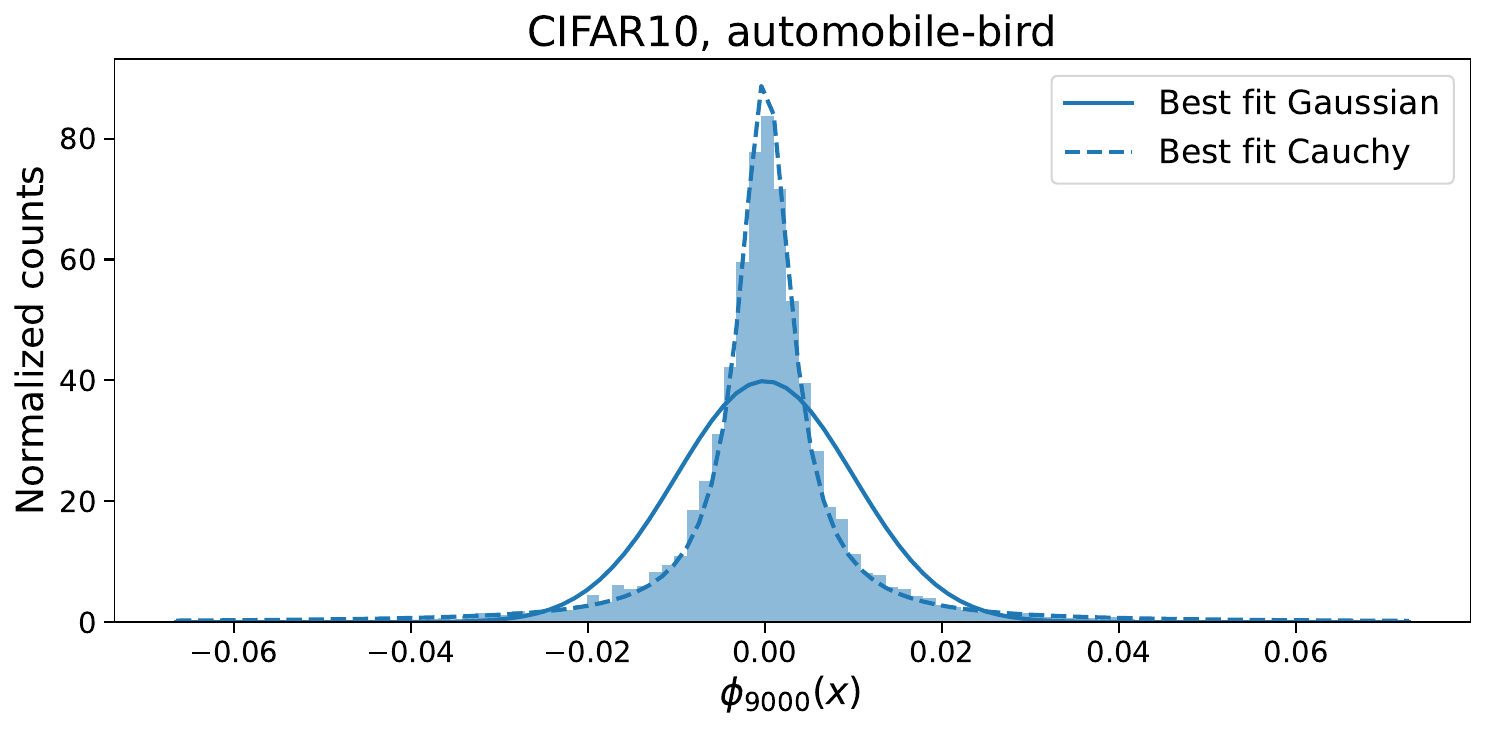}
    \caption{ \textbf{Example distribution of high-$k$ feature modes.} By comparing with figure~\ref{fig:empirical_results_scalinglaws_gaussianfeaturesproof}, we can see that the feature modes shift from being best described by a Gaussian distribution to being better described by a Cauchy distribution for high $k$ values. The best fit Gaussian and best fit Cauchy distributions are overlaid.}  
    \label{fig:cauchy_example}
\end{figure}

In figure~\ref{fig:empirical_results_scalinglaws_gaussianfeaturesproof} we saw that the feature modes for real-world datasets, like the ones considered in section~\ref{sec:scalingLaws_empirical} (MNIST and CIFAR10), are often Gaussian distributed. This interesting fact has been previously asserted in the literature~\cite{Simon2021}, motivating Gaussianity as a starting point for theoretical descriptions. In this appendix, we explore the dependence of this statement as a function of $k$ for the real-world datasets considered in this work. Namely, we can see in figure~\ref{fig:cauchy_example} that as we go to very high $k$ values, the distribution of feature modes become better described by a Cauchy distribution.

\begin{figure}
    \centering
    \includegraphics[width=0.49\linewidth]{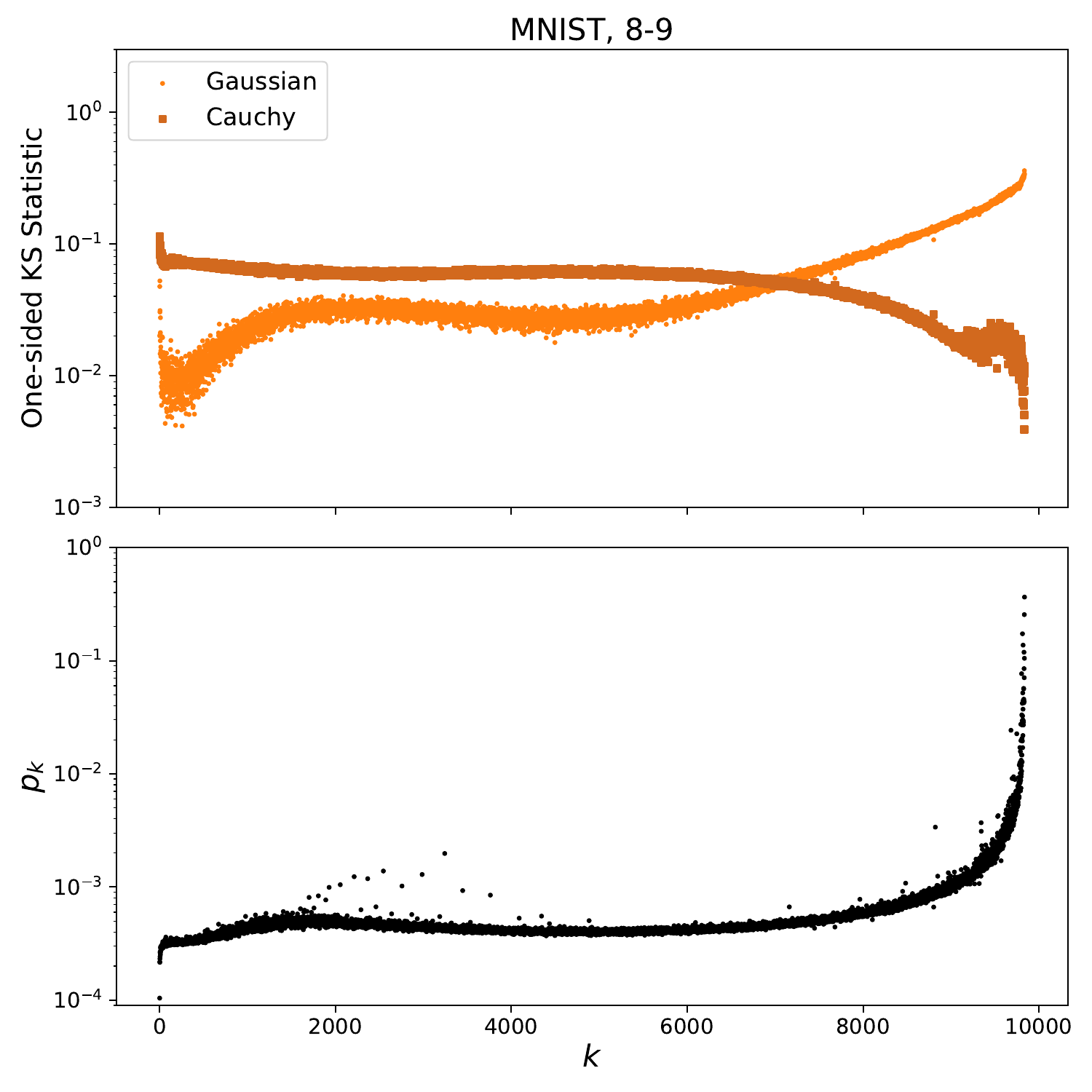}
    \includegraphics[width=0.49\linewidth]{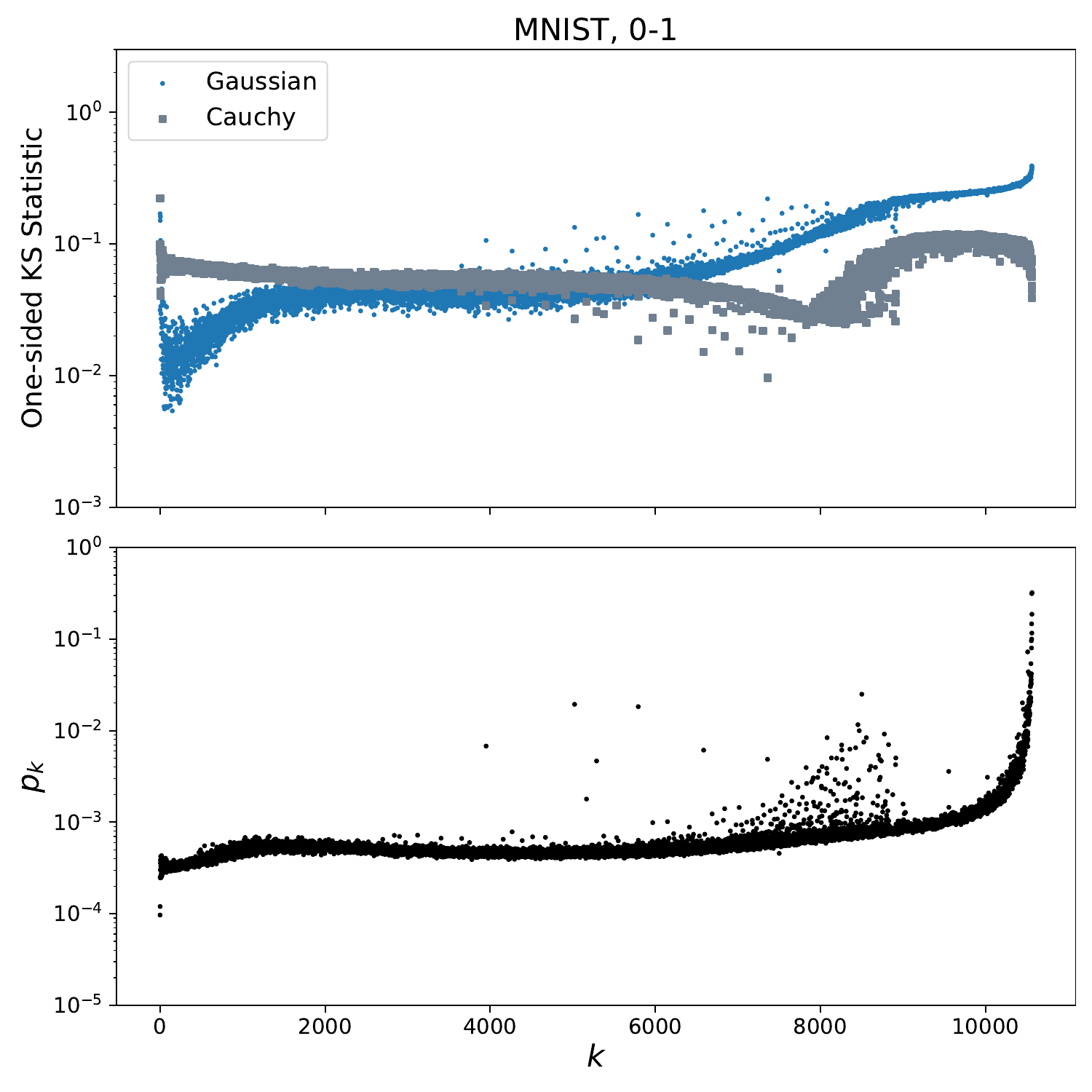}
    \includegraphics[width=0.49\linewidth]{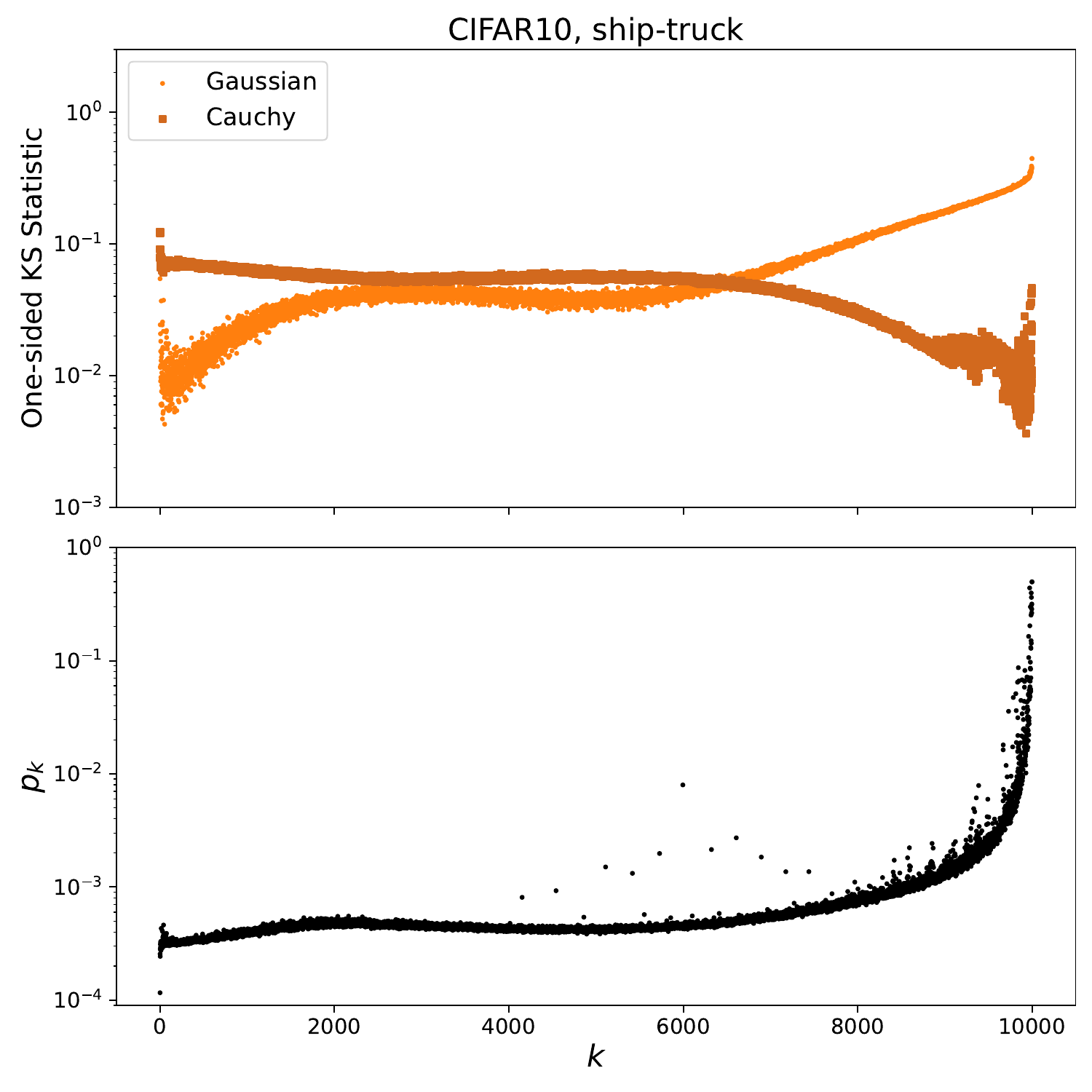}
    \includegraphics[width=0.49\linewidth]{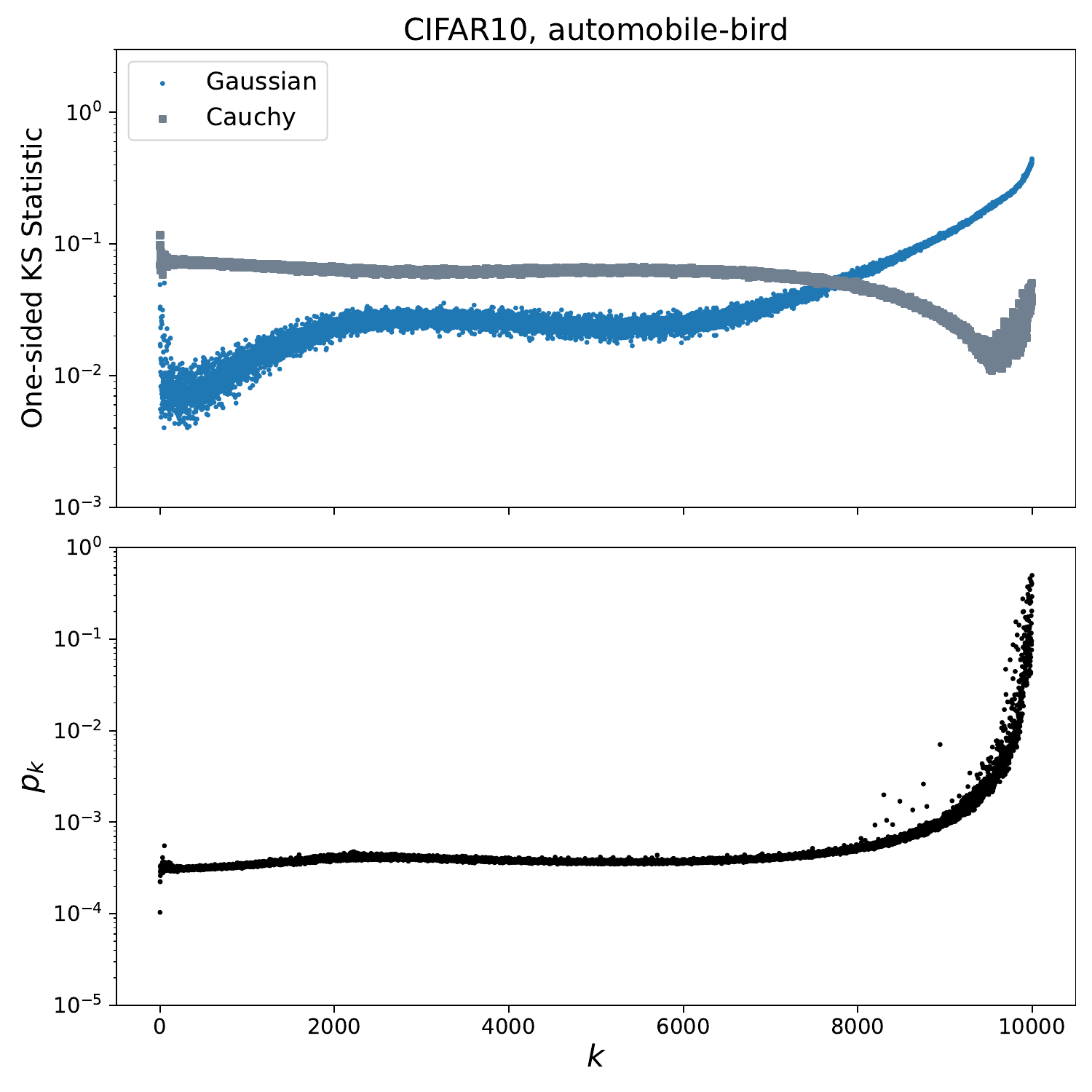}
    \caption{ \textbf{Distribution of feature modes as a function of $k$.} Here we show that feature modes shift from being best described by a Gaussian distribution to being better described by a Cauchy distribution at high $k$ values. For each mode, we perform a fit to both a Gaussian and Cauchy distribution and plot the one-sided KS test statistic between the empirical data and best fit as a function of $k$. Indeed, for very high $k$, we can see that a Cauchy distribution becomes a better description. By also showing the inverse participation ratio, $p_k$~\cite{THOULESS197493}, we see that these Cauchy-like modes correspond to more data-point-localized features (as indicated by having a larger inverse participation ratio).}    
    \label{fig:cauchy_vs_gaussian}
\end{figure}

Figure~\ref{fig:cauchy_vs_gaussian} further explores this trend; for each $k$ we perform both a Gaussian and Cauchy fit to the feature mode distribution and analyze the quality of the fit by calculating the one-sided KS test statistic between the empirical data and best fit. We note that for the majority of the $k$ range, a Gaussian distribution remains a better approximation, as indicated by having a lower KS test statistic value. We additionally calculate the inverse participation ratio, $p_k$~\cite{THOULESS197493}, which is defined as 
\begin{equation}
    p_k := \sum_i \frac{[\varphi_k(x_i)]^4}{\left(\sum_j [\varphi_k(x_j)]^2\right)^2}~.
\end{equation}
We can see that these Cauchy-like modes  correspond to more data-point-localized features (as indicated by having a larger $p_k$). Moreover, these Cauchy-like modes describe well the bulk of the probability for very high $k$ which corresponds to very small $\lambda_k$. Thus, the contribution to $\sigma_{\rm eff}^2$ from these very high ($k>8000$) modes is $\mathcal{O}(1\%)$. Overall, we can see that a Gaussian approximation of feature modes is well-motivated but still incomplete; the presence of such non-Gaussian features motivates studying corrections to this approximation, as has been done in this work (see section~\ref{sec:beyond}.) 

\vfill

\bibliographystyle{ytphys}
\bibliography{biblio}

\end{document}